\documentclass[final]{nesy2026} 

\usepackage{mathtools}
\usepackage{booktabs} 
\usepackage{multirow} 
\usepackage{caption}
\usepackage{subcaption} 
\usepackage{enumitem}
\usepackage{csquotes}
\usepackage[table]{xcolor}
\usepackage{graphbox}
\usepackage{booktabs}
\usepackage{multirow}
\usepackage{graphicx}
\usepackage{pgfplots}
\usepackage[table]{xcolor}
\pgfplotsset{width=10cm,compat=1.9}
\usepgfplotslibrary{external}
\usepackage[capitalize]{cleveref}
\crefname{section}{Sec.}{Secs.}
\Crefname{section}{Section}{Sections}
\Crefname{table}{Table}{Tables}
\crefname{table}{Tab.}{Tabs.}

\let\cite\citep

\makeatletter
\newcommand\footnoteref[1]{\protected@xdef\@thefnmark{\ref{#1}}\@footnotemark}
\makeatother

\begin{document}

\title[Weakly Supervised Concept Learning]{Weakly supervised concept Bottleneck Learning for Robust Two stage Object centric visual reasoning}

\author{%
  \Name{Sparsh Tiwari}\\
  \addr University of L\"ubeck, Germany
  \AND
  \Name{Gesina Schwalbe} \\
  \addr Ulm University, Germany
  \AND
  \Name{Bettina Finzel} \\
  \addr University of Bamberg, Germany
}

\maketitle

\begin{abstract}
Two-stage neuro-symbolic architectures provide an elegant paradigm for visual problem solving by cleanly separating connectionist perception of predefined symbols from possibly later defined relational reasoning thereon. 
However, anchoring high-level predicates into visual frames typically necessitates 
annotations that are 
expensive to acquire. In this work, we introduce the Dynamic Orthogonal Concept Bottleneck (D-OCB), an object-centric slot-VAE framework designed to extract 
human-aligned symbolic predicates under extremely weak supervision. D-OCB eliminates the arduous manual tuning of loss-balancing coefficients by dynamically learning optimal hyperparameter allocations during training. To infuse prior knowledge on independence of concept categories, in addition to standard reconstruction self-supervision we penalize correlation across concept subspaces. Crucially, to combat the instability of very low supervision regimes, D-OCB incorporates a dynamic dimensionality allocation mechanism; this adaptive formulation allows well-represented concepts to yield latent dimensions to underperforming concepts that are lagging behind, effectively preventing representation collapse and significantly improving overall concept accuracy. Through an extensive empirical evaluation, we demonstrate that our framework achieves high concept alignment and downstream visual reasoning accuracy using minimal label budgets, matching or outperforming end-to-end paradigms.

\end{abstract}

\section{Introduction}

Neuro-symbolic architectures for visual tasks allow to combine the feature extraction capabilities of deep neural networks (DNNs) on raw signals with interpretable and efficient symbolic reasoning \cite{garcez2023neurosymbolic, Kishor2022NeuroSymbolicAB}. 
A central impediment within this paradigm is the classic symbol grounding problem, which asks how abstract symbols acquire semantic meaning beyond circular, text-based definitions and connect to perceptual boundaries in the physical world \cite{Minsky1991LogicalVA}.
Specifically, feature extractors have to aggregate
distributed features across visual space 
into coherent, unified object-centric representations \cite{brady2024interactionasymmetrygeneralprinciple}. Two-stage neuro-symbolic architectures offer a compelling solution when symbol annotations are available, by structurally decoupling perception of symbols from reasoning on them \cite{mao2018neurosymbolic, manhaeve2018deepproblogneuralprobabilisticlogic}. In the first stage, a perception network maps raw inputs onto an explicit concept bottleneck layer; in the second stage, a symbolic engine executes logic programs directly over these discovered predicates. This modularity ensures that predictions are fully traceable, reduces susceptibility to shortcut learning, yields faster optimization convergence, and permits the reuse of a static reasoning backend across disparate visual domains \cite{kikaj2025deepgraphlog}.

Despite these advantages, traditional concept bottlenecks suffer from labeling requirements. 
Existing frameworks rely on full supervision to instantiate intermediate concepts, or 
turn to reinforcement learning settings with high sample variance \cite{dai2019diagnosingenhancingvaemodels}. The use of background knowledge or programmatic constraints to provide weak supervision, where only parts of the training data requires costly labels, remains significantly underexplored \cite{knab2026whats}.
This particularly is the case for object-centric concept bottleneck architectures and for the use in neuro-symbolic architectures.

\begin{figure}
    \centering
    \vspace*{-2\baselineskip}
    \includegraphics[width=0.75\linewidth]{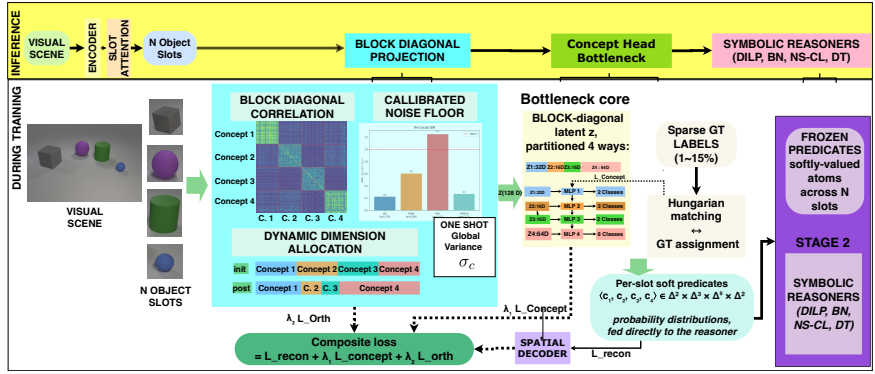}
    \vspace*{-.5\baselineskip}
    \caption{Overview of our D-OCB architecture and training scheme}
    \label{fig:visual-abstract}
    \vspace*{-\baselineskip}
\end{figure}

To effectively utilize weak supervision for object-centric extraction, three concrete representational hurdles must first be resolved. First, generative bottlenecks trained with sparse labels frequently succumb to \textbf{representation collapse} \cite{DBLP:journals/corr/abs-2110-09348}, where unbounded variance optimization forces latent variables to degrade into uninformative noise. Second, without exhaustive dense annotations to explicitly isolate attributes, feature extractors naturally default to \textbf{feature cross-correlation}. This results in the entanglement of distinct conceptual properties (e.g., blending shape constraints into color representations) to exploit spurious visual shortcuts. Finally, traditional bottlenecks assign statically fixed vector sizes across all target concepts. This assumes uniform semantic complexity and inevitably causes \textbf{capacity misallocation} \cite{Hou2022}, where highly nuanced concepts are starved of representational bandwidth while trivial concepts waste feature dimensions.

To address these vulnerabilities, we introduce the Dynamic Orthogonal Concept Bottleneck (D-OCB \footnote{\tiny The codebase is available at https://github.com/sparshX1993/Dynamic-Orthogonal-Concept-Bottleneck.}), illustrated in Figure \ref{fig:visual-abstract}. Rather than serving as a standalone, end-to-end reasoning architecture, D-OCB is explicitly framed as a weakly supervised predicate-grounding module (or perceptual bottleneck) designed specifically for two-stage neuro-symbolic pipelines. By structurally uncoupling perception from deduction, D-OCB anchors human-aligned symbolic predicates from minimal annotation budgets. It prevents representation collapse by anchoring per-concept variances against a calibrated noise floor alongside a variational auto-encoder reconstruction loss~\cite{sutter2025unitydiversityimprovedrepresentation}. To combat feature cross-correlation, D-OCB introduces an explicit penalty that orthogonalizes subspaces between different concept categories. Crucially, to prevent capacity misallocation , D-OCB utilizes a novel dynamic dimensionality allocation mechanism that adaptively redistributes feature space size during training based on concept convergence; this ensures stable representations before matching concepts via the Hungarian algorithm. The result is a specialized perception module capable of providing crisp, human-aligned predicate distributions to any frozen, off-the-shelf symbolic solver.
Our key \textbf{contributions} amount to:
  %
  %
  %
%
\begin{itemize}[nosep,leftmargin=1em]
  \item We introduce D-OCB (cf.\,\autoref{fig:visual-abstract}), a weakly supervised, two-stage object-centric visual reasoning framework that grounds semantic predicates using as little as $1\%$ of instance-level concept annotations.
  \item We propose a structural approach for stable concept learning under weak supervision that effectively prevents rank collapse inside each concept block.
  \item We formalize a dynamic dimensionality allocation protocol, entirely eliminating the need for manual latent-capacity tuning.
  \item We empirically demonstrate on the CLEVR and confounded CLEVR-Hans3/7 benchmarks that D-OCB predicates drive independent symbolic reasoners to accuracies competitive with end-to-end differentiable paradigms, adding as little as $1\%$-$15\%$ concept supervision to the training data.
\end{itemize}

\section{Related Work}
\label{sec:related_work}
Our architecture is fundamentally motivated by the need to solve two classical cognitive hurdles simultaneously: the symbol grounding problem \cite{Minsky1991LogicalVA}, which asks how discrete symbols acquire physical semantic meaning from unstructured visual boundaries; and the binding problem \cite{brady2024interactionasymmetrygeneralprinciple}, which concerns how distributed features across visual space are aggregated into unified, coherent entities.
To achieve this, the design of our D-OCB intersects three major paradigms in contemporary neuro-symbolic artificial intelligence: concept bottleneck models (CBMs), object-centric representation learning, and differentiable logical rule induction.

\textbf{CBMs and Generative Realignment.}
CBMs are interpretable architectures that factor downstream predictions through an intermediate bottleneck layer, where each individual output dimension is explicitly constrained to represent exactly one human-understandable concept\cite{knab2026whats,kazhdan2021disentanglement}. Traditional formulations in this domain typically operate over holistic, image-level encodings and rely on dense, exhaustive supervision to map pixels onto categorical concepts \cite{knab2026whats,losch2021semantic}. To mitigate this data-scarcity issue \cite{agrawal2025walkingwebconceptclassrelationships}, contemporary frameworks have introduced semi-supervised \cite{Hu_2025_ICCV} and completely label-free \cite{oikarinen2023labelfree} concept learning. Concurrently, efforts have been made to move beyond holistic representations by developing object-centric concept bottlenecks \cite{NEURIPS2025_639d992f}. Alongside these developments, recent frameworks have explored the integration of unsupervised generative constraints into the bottleneck layer \cite{locatello2019challenging}. These generative approaches formalize how an image reconstruction loss can actively optimize concept alignment, prevent information leakage, and isolate target attributes \cite{kim2018disentangling,locatello2019fairness}. D-OCB builds directly upon these generative and object-centric foundations but shifts the paradigm entirely to extremely sparse supervision regimes. By introducing a tuning-free dynamic allocation scheme and an explicit cross-correlation constraint, our work explores how balancing self-supervised reconstruction with minimal concept labels can preserve or even enhance grounding accuracy \cite{chen2019isolatingsourcesdisentanglementvariational}.

\textbf{Object-Centric Tokenization and Slot-Based Bottlenecks.}
To perform relational and compositional reasoning in multi-object scenes, perception models require an object-centric tokenization \cite{brady2024interactionasymmetrygeneralprinciple, seitzer2023bridginggaprealworldobjectcentric}. A recent and superior approach is to use slot attention mechanisms which decompose a raw visual frame into distinct, non-overlapping slot tokens \cite{didolkar2025ctrlolanguagecontrollableobjectcentricvisual}, encouraging spatial specialization and compositionality \cite{locatello2020objectcentric}. Despite their success in unsupervised scene decomposition \cite{mosbach2025soldslotobjectcentriclatent}, extending object-centric slots to interpretable concept layers introduces significant training instabilities \cite{liu2025metaslotbreakfixednumber}. D-OCB resolves these limitations by introducing a strictly block-diagonal projection matrix as additional bias that forces uncorrelated concept subspaces.

\textbf{Neuro-Symbolic Reasoning and Two-Stage Pipelines.}
Neuro-symbolic AI seeks to combine the robust perceptual pattern recognition of deep neural networks with the rigorous deductive compositionality of symbolic reasoning backends \cite{shindo2021neurosymbolicforwardreasoning,roth2025enhancing}.
While end-to-end logical frameworks like differentiable inductive logic programming (DILP) \cite{kikaj2025deepgraphlog,shindo2023ailp,DBLP:journals/corr/abs-2103-01719} optimize perception and deduction jointly via gradient descent, they remain highly vulnerable to Likelihood Blow-up \cite{mattei2018leveraging}  and Posterior Collapse and representation contamination. To make use of available ground truth concept labels for interpretability, and to enable systematic reuse, D-OCB operates as a modular, two-stage neuro-symbolic framework that structurally uncouples perceptual tokenization from downstream symbolic solver engines \cite{dinu2024symbolicaiframeworklogicbasedapproaches,wüst2025synthesizingvisualconceptsvisionlanguage}.

\section{Approach: Dynamic Orthogonal Concept Bottleneck}
\label{sec:approach}

End-to-end differentiable reasoning frameworks are highly susceptible to feature entanglement \cite{marconato2023neurosymbolicconceptscreatedequal} and rely heavily on dense supervision. To overcome these inherent limitations, we propose a two-stage neuro-symbolic architecture that cleanly decouples object-centric visual perception from downstream logical reasoning. To ensure the perception module grounds continuous visual data into discrete, disentangled symbols under extreme weak supervision ($1\%$--$15\%$), we introduce D-OCB. 

For a formal task definition, let $\mathcal{X} \in \mathbb{R}^{H \times W \times C}$ define the input space of raw visual scenes, and let $\mathcal{S} = (s_1, \dots, s_K)$ represent an ordered specification of human-interpretable concept predicates. Our goal is to train an object-centric perception module $g: \mathcal{X} \rightarrow \mathcal{C}$ that maps an input image $x$ to a discrete valuation tensor $c \in \mathbb{R}^{N \times K}$, representing the presence of $K$ concepts across $N$ distinct object slots, utilizing an extremely limited label budget. These grounded concepts are then passed into a standalone symbolic solver $f: \mathcal{C} \rightarrow \mathcal{Y}$ to deduce the final downstream task solution $y \in \mathcal{Y}$ \cite{muggleton1991inductive}.

\subsection{Overview: Architecture and Optimization Objectives}

We first establish the overall architecture for object-centric perceptual grounding, mapping raw pixels to discrete symbolic predicates in a step-by-step manner.An input scene $x \in \mathbb{R}^{h \times w \times c}$ of height $h$, width $w$, and $c$ channels is processed by a feature extractor, such as DINOv2 \cite{oquab2024dinov2learningrobustvisual}, augmented with positional embeddings. The resulting continuous feature maps are routed through an iterative slot attention module \cite{locatello2020objectcentric} to enforce a tokenization prior, outputting $N$ uninterpretable object slots:
\begin{gather}
H = \text{SlotAttention}(\text{Encoder}(x)) \in \mathbb{R}^{N \times D}
\end{gather}
where $D$ represents the hidden slot dimension (typically $D=128$).

Following the extraction of these deterministic slots, the model maps the representations to a partitioned latent concept space $Z \in \mathbb{R}^{N \times D_{\text{total}}}$. This concept space has a total feature dimensionality $D_{\text{total}}$, which is explicitly divided into independent subspaces for each target concept $c \in \mathcal{C}$ (e.g., shape, color), such that $D_{\text{total}} = \sum_{c \in \mathcal{C}} D_c$. The specific dimensionality $D_c$ allocated to each concept is not static; it is dynamically adjusted during training (as detailed in Section \ref{sec:Ortho}) to provide more capacity to struggling concepts.

To derive crisp concept predictions from these latent representations, each isolated concept subspace $z_c \in \mathbb{R}^{D_c}$ per slot is fed into an independent prediction head (a 2-layer MLP). These heads output raw class logits, which are converted into continuous probabilities via a softmax function (for multi-class concepts like color) or a sigmoid function (for binary concepts like material/Size). Finally, an $\arg\max$ operation (or standard thresholding) discretizes these soft probabilities into the definitive, crisp symbolic predicates (e.g., \texttt{color(obj1, red)}) that are ultimately passed to the downstream logical reasoner.

To train this bottleneck effectively without manual hyperparameter engineering, we optimize an object-centric Variational Autoencoder (VAE) objective that balances a self-supervised reconstruction loss $\mathcal{L}_{\text{recon}}$, a concept prediction loss \cite{zbontar2021barlow} $\mathcal{L}_{\text{concept}}$, and a structural cross-correlation loss\footnote{\tiny For more detailed breakdown of our loss function  and its mathematical formulations, see supplementary material.}  \cite{bardes2022vicreg} $\mathcal{L}_{\text{orth}}$:
\begin{gather}
\SwapAboveDisplaySkip
\mathcal{L}_{\text{total}} = \mathcal{L}_{\text{recon}} + \lambda_{1} \mathcal{L}_{\text{concept}} + \lambda_{2} \mathcal{L}_{\text{orth}}
\end{gather}
The scaling coefficients $\lambda_1$ and $\lambda_2$ are dynamically updated at each epoch based on the moving average of gradient variances. Crucially, the cross-correlation loss $\mathcal{L}_{\text{orth}}$ minimizes the off-diagonal elements of the feature correlation matrix. This explicitly forces the learned concept subspaces to remain orthogonal and causally isolated from one another \cite{9363924, suter2019robustlydisentangledcausalmechanisms}, ensuring that weakly supervised concept signals are not overwhelmed by unconstrained reconstruction errors \cite{watters2019spatialbroadcastdecodersimple} and avoiding the feature interference common in standard dense projections \cite{goyal2022inductivebiasesdeeplearning, kirsch2018modular}.


\subsection{Generative Bottleneck: Calibrated Fixed Noise Floors}

A critical challenge in this architecture is the representation collapse common to generative bottlenecks \cite{heek2026unifiedlatentsultrain}. Generative models typically learn a per-sample variance ($\sigma(x)$). When annotation is scarce, the unsupervised reconstruction loss $\mathcal{L}_{\text{recon}}$ can force the network to map visual noise directly into this learned variance. This unbounded optimization causes a likelihood blow-up \cite{mattei2018leveraging}, which leads to severe posterior collapse \cite{lucas2019dont}. Consequently, the latent variables become mathematically non-identifiable \cite{wang2023posterior}, effectively destroying the semantic manifold. 

To prevent this, D-OCB discards per-sample learned variance entirely. Instead, we introduce a fixed noise floor $\sigma_c$ specific to each concept subspace $c$:
\begin{gather}
\SwapAboveDisplaySkip
z_c = \mu_c(x) + \sigma_c \odot \epsilon\end{gather}

Optimized globally during a brief one-shot calibration phase on the labeled subset, $\sigma_c$ is permanently frozen. This renders the signal-to-noise ratio ($\text{SNR} = ||\mu_c||^2 / \sigma_c^2$) mathematically stable and bars the decoder from encoding reconstruction noise into the variance.

\subsection{Orthogonality: Concept Assignment and Dimensionality Allocation}
\label{sec:Ortho}

With the latent space cleanly orthogonalized by the cross-correlation loss and stabilized by the fixed noise floors, the remaining task is to map these dimensions to specific human-interpretable concepts and distribute capacity effectively. 

First, to explicitly assign specific latent dimensions to their corresponding ground-truth concepts (e.g., matching a subspace to concept categories ``shape'' or ``color''), alignments are secured via the Hungarian matching algorithm \cite{Kuhn1955TheHM}.

Second, to efficiently allocate latent capacity among these matched concepts without rank collapse \cite{noci2022signalpropagationtransformerstheoretical}, D-OCB replaces standard Singular Value Decomposition \cite{DBLP:journals/corr/abs-1710-02812} with Fisher's Linear Discriminant Analysis \cite{articlechen} (LDA). LDA projects running class centroids by computing a projection matrix $W\coloneqq \arg\max_W \tfrac{|W^T S_B W|}{|W^T S_W W|}$ that explicitly maximizes between-class scatter $S_B$ relative to within-class scatter $S_W$.
%
By maximizing class separability, LDA remains invariant to unsupervised reconstruction noise. Combined with Johnson-Lindenstrauss random projections for distance-preserving weight transfers, this dynamic dimensionality allocation ensures that underperforming concepts receive the necessary representational capacity without the need for manual latent-capacity tuning.

\begin{figure}
    \centering
    \vspace*{-1cm}%
    \includegraphics[width=0.3\linewidth]{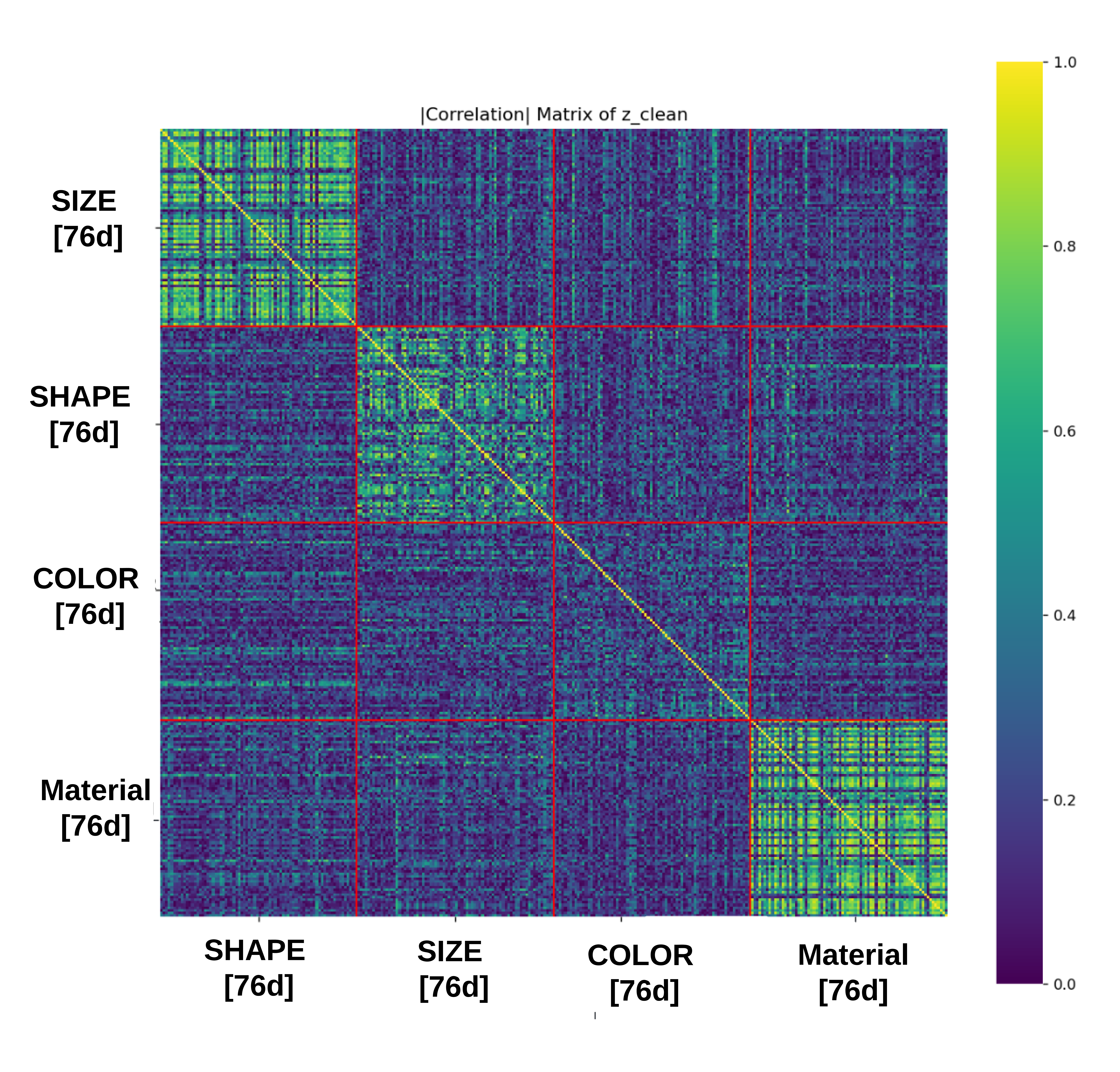}
    \includegraphics[width=0.3\linewidth]{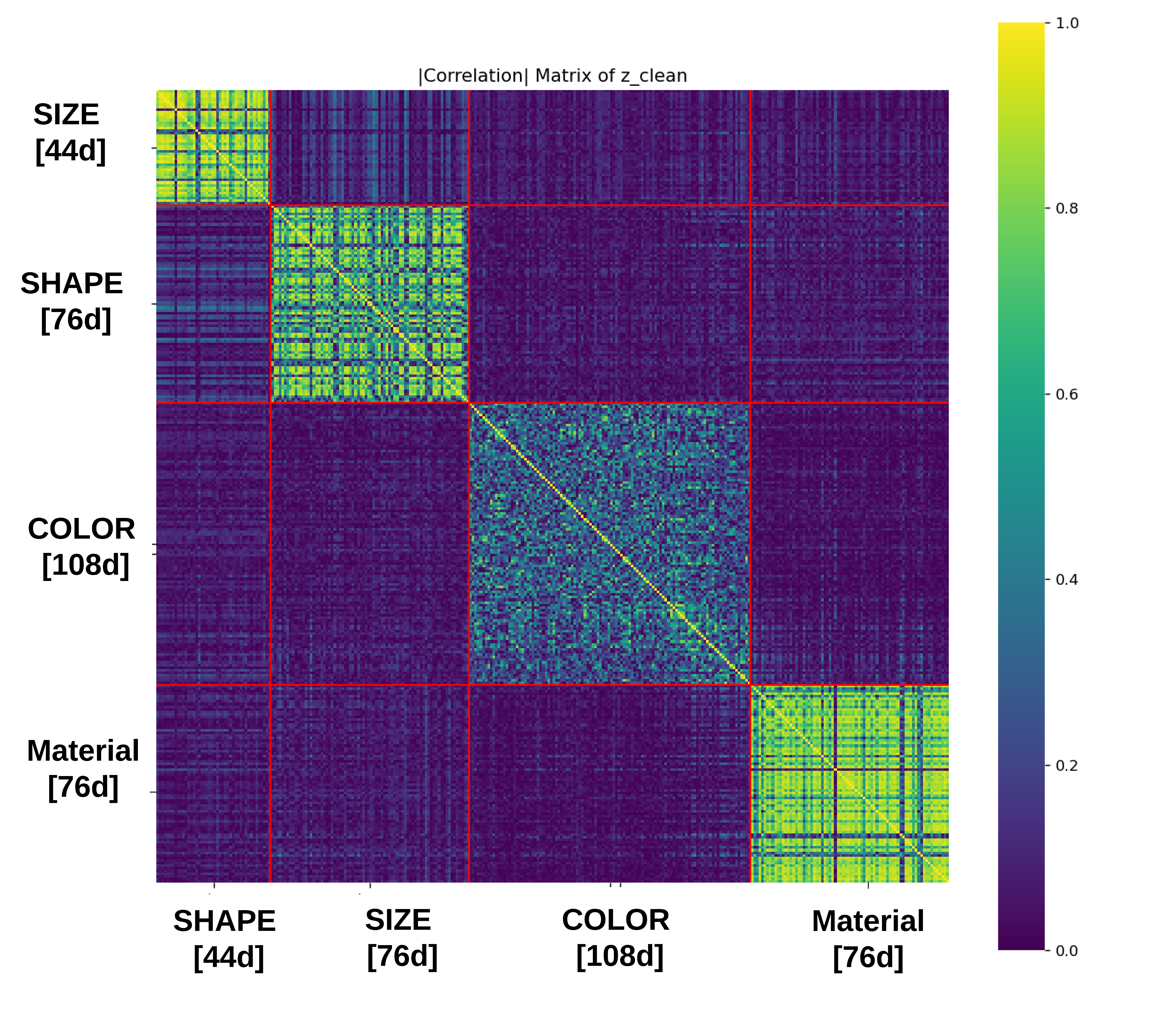}
    \vspace*{-.5cm}%
    \caption{\small Correleation MATRIX for Clevr HANS7 Showing our DDA assign dimension from one concept to another for epoch 5 (left) to epoch 100 (right)}
    \label{fig:correlation}
\end{figure}

\section{Experiments and Results}

We empirically evaluate the representational fidelity, reasoning capability, and data efficiency of D-OCB, aiming to answer the following questions:

\noindent\textbf{RQ1:} Can D-OCB extract highly accurate and well-grounded visual predicates under extreme weak supervision (using only $1\%$--$15\%$ of the training dataset annotations)?

\noindent\textbf{RQ2:} Are the extracted predicates sufficiently stable to drive an offline symbolic solver without requiring joint end-to-end logic backpropagation?

\noindent\textbf{RQ3:} Does structurally decoupling perception from logical reasoning successfully prevent the exploitation of spurious correlations, thereby enabling D-OCB to achieve robust out-of-distribution (OOD) confounder resistance?

\paragraph{Datasets.}
We evaluate on the CLEVR object-attribute benchmark (70,000 training images) \cite{johnson2016clevrdiagnosticdatasetcompositional} and its logical scene classification variants CLEVR-Hans3 (3,000 per class, 3 classes) and CLEVR-Hans7 (21,000 training images) \cite{DBLP:journals/corr/abs-2011-12854} and a real world melanoma detection application using the HAM10000 dataset \footnote{find details on the choice of datasets in the technical supplementary} \cite{DBLP:journals/corr/abs-1803-10417} (8,012 training images).
Hence, the considered sample budgets for 1\%/15\% weak supervision amount to a mere $700$/$10.3k$, $90$/$1.35k$, $210$/$3.15k$, $80$/$1.2k$ samples for CLEVR, CLEVR3, CLEVR7, and HAM10000 respectively.
In all weakly supervised configurations, the remaining majority of the training pool remains completely unannotated. 

\paragraph{Baselines.}
We benchmark our semantic tokenization pipeline against the following contemporary visual concept extraction architectures: the Concept Embedding Model (CEM) \cite{zarlenga2022conceptembeddingmodelsaccuracyexplainability}; the Probabilistic Concept Bottleneck Model (ProbCBM) \cite{kim2023probabilisticconceptbottleneckmodels}; SlotFormer \cite{biza2023invariantslotattentionobject} for temporal object-centric mechanics; a Standard Slot Attention baseline (Vanilla+SA); SlotVAE-NO-KL (a deterministic ablation); and SlotVAE-KL \cite{wang2024slotvaeobjectcentricscenegeneration}, optimized under standard Kullback-Leibler regularizers \cite{asperti2020balancingreconstructionerrorkullbackleibler, wang2023posteriorcollapselatentvariable, lucas2019dontblameelbolinear}.To rigorously evaluate out-of-distribution (OOD) reasoning and confounder resistance, we compare our modular D-OCB (coupled with DILP) against end-to-end and dense-supervision paradigms. These include: an end-to-end trained CNN (0\% concept supervision); a CNN\,+\,DILP baseline (100\% supervision); state-of-the-art end-to-end differentiable frameworks ($\alpha$ILP \cite{shindo2023ailp} and NEUMANN  \cite{shindo2023learning}); DeepProbLog \cite{ manhaeve2018deepproblogneuralprobabilisticlogic} (trained jointly with 100\% supervision); and a Ground Truth (Oracle) baseline.

\paragraph{Evaluation Metrics.}

Perceptual quality and grounding alignment are assessed via concept accuracy (acc.) and Macro-F1 scores, computed by aligning predicted object slots to ground-truth annotations via Hungarian matching \cite{Kuhn1955TheHM}. Downstream task execution is measured via rule evaluation acc. and F1-scores \cite{daniele2024simpleeffectivetransferlearning}.

\paragraph{Hyperparameters.}
The architecture is optimized over 200 epochs with a batch size of 64 and a base learning rate of $4 \times 10^{-4}$ utilizing a cosine annealing schedule \cite{oquab2024dinov2learningrobustvisual}. DDA capacity shifts occur every 5 epochs following an initial 30-epoch warmup gate, transferring features in discrete steps of $\Delta = 8$ dimensions to maintain block-diagonal alignment. To prioritize semantic grounding, the concept classification loss is heavily weighted ($\lambda_1  = 8.0$, scaling to $12.0$ in later phases) relative to the reconstruction baseline ($\lambda_{recon} = 1.0$). Finally, the object existence head employs a focal loss ($\gamma = 2.0$, positive weight $= 2.5$) to successfully counter slot sparsity.


\subsection{RQ1: Visual Concept Extraction and Predicate Accuracy}
\label{sec:rq1}

To evaluate the representational quality of our perception module, we measure its capability to ground continuous visual features into discrete symbolic predicates using the CLEVR and HAM10000 datasets under extremely weak supervision regimes ($1\%$ to $15\%$). Unlike traditional slot models that frequently exhibit performance collapse on complex attributes due to unconstrained reconstruction objectives hijacking the latent space topology, our D-OCB explicitly prevents features from dissolving into unstructured noise.

\begin{table*}[t]
\centering
\vspace*{-2\baselineskip}
\caption{\small Concept classification acc.\ on the CLEVR dataset across varying levels of weak supervision. Results are reported as mean $\pm$ standard deviation over 5 folds.}
\vspace{-3mm}
\label{tab:rq1_clevr_updated}
\resizebox{\textwidth}{!}{\tiny%
\begin{tabular}{@{}l@{\,}l@{~} c@{~}c@{~}c@{~}c@{~}c @{~~}c@{~~} c@{~}c@{~}c@{~}c@{~}c@{}}
\toprule
& & \multicolumn{5}{c}{\textbf{CNN Backbone}} && \multicolumn{5}{c}{\textbf{DINO Backbone}} \\
\cmidrule{3-7} \cmidrule{9-13}
\textbf{Sup.} & \textbf{Method} & Size & Shape & Color & Material & Mean && Size & Shape & Color & Material & Mean \\
\midrule
\multirow{7}{*}{\textbf{1\%}} & Vanilla+SA & $51.2 \pm 0.0$ & $49.3 \pm 0.0$ & $10.7 \pm 0.0$ & $62.2 \pm 0.0$ & $46.4 \pm 0.0$ && $72.3 \pm 0.0$ & $63.4 \pm 0.0$ & $15.8 \pm 0.0$ & $70.6 \pm 0.0$ & $57.3 \pm 0.0$ \\
 & CEM & $51.2 \pm 1.3$ & $33.9 \pm 0.5$ & $17.2 \pm 6.3$ & $50.9 \pm 0.5$ & $38.3 \pm 2.1$ && $73.4 \pm 2.2$ & $41.3 \pm 3.9$ & $12.3 \pm 0.1$ & $70.8 \pm 6.6$ & $49.5 \pm 2.6$ \\
 & ProbCBM & $48.2 \pm 3.9$ & $34.5 \pm 2.1$ & $20.9 \pm 8.2$ & $50.7 \pm 0.7$ & $38.6 \pm 3.3$ && $72.3 \pm 4.4$ & $38.4 \pm 2.2$ & $12.5 \pm 0.3$ & $67.0 \pm 7.4$ & $47.5 \pm 3.2$ \\
 & SlotFormer & $25.2 \pm 3.1$ & $25.0 \pm 2.0$ & $17.5 \pm 5.1$ & $50.1 \pm 1.1$ & $29.4 \pm 1.7$ && $48.0 \pm 2.8$ & $38.3 \pm 2.8$ & $17.7 \pm 3.9$ & $55.0 \pm 3.1$ & $39.8 \pm 2.1$ \\
 & SlotVAE-NoVAE & $52.8 \pm 0.0$ & $51.5 \pm 0.0$ & $8.9 \pm 0.0$ & $60.5 \pm 0.0$ & $48.6 \pm 0.0$ && $76.2 \pm 0.0$ & $62.6 \pm 0.0$ & $12.1 \pm 0.0$ & $75.6 \pm 0.0$ & $56.9 \pm 0.0$ \\
 & SlotVAE-KL & $62.0 \pm 4.8$ & $64.0 \pm 6.6$ & $16.9 \pm 3.7$ & $79.2 \pm 1.4$ & $55.0 \pm 3.2$ && $90.5 \pm 6.7$ & $80.7 \pm 6.4$ & \textbf{38.3 $\pm$ 0.2} & $93.6 \pm 3.2$ & $75.3 \pm 3.3$ \\
 & \textbf{D-OCB (Ours)} & \textbf{75.1 $\pm$ 16.1} & \textbf{67.5 $\pm$ 3.4} & \textbf{28.2 $\pm$ 5.1} & \textbf{84.5 $\pm$ 5.0} & \textbf{63.8 $\pm$ 7.1} && \textbf{94.4 $\pm$ 1.4} & \textbf{82.3 $\pm$ 8.6} & $36.7 \pm 6.3$ & \textbf{94.9 $\pm$ 2.3} & \textbf{77.1 $\pm$ 4.5} \\
\midrule
\multirow{5}{*}{\textbf{5\%}} & CEM & $53.5 \pm 1.1$ & $35.9 \pm 0.6$ & $24.5 \pm 0.5$ & $51.9 \pm 1.0$ & $41.5 \pm 0.5$ && $80.2 \pm 2.9$ & $48.7 \pm 1.5$ & $12.4 \pm 0.2$ & $82.7 \pm 2.9$ & $56.0 \pm 1.0$ \\
 & ProbCBM & $48.1 \pm 5.2$ & $34.5 \pm 1.5$ & $25.0 \pm 0.9$ & $50.9 \pm 0.7$ & $39.6 \pm 1.6$ && $74.6 \pm 5.5$ & $46.8 \pm 4.9$ & $12.6 \pm 0.4$ & $79.8 \pm 3.9$ & $53.5 \pm 3.1$ \\
 & SlotFormer & $26.7 \pm 2.7$ & $26.1 \pm 3.2$ & $24.8 \pm 0.5$ & $50.3 \pm 0.8$ & $32.0 \pm 0.9$ && $46.4 \pm 3.1$ & $43.9 \pm 4.1$ & $14.1 \pm 1.8$ & $63.2 \pm 3.9$ & $41.9 \pm 2.4$ \\
 & SlotVAE-KL & $85.8 \pm 5.2$ & $68.5 \pm 4.5$ & $43.0 \pm 2.3$ & \textbf{91.9 $\pm$ 1.6} & $72.3 \pm 2.6$ && $91.0 \pm 7.9$ & $87.3 \pm 5.9$ & $62.3 \pm 0.1$ & $96.6 \pm 5.6$ & $84.1 \pm 4.1$ \\
 & \textbf{D-OCB (Ours)} & \textbf{91.8 $\pm$ 3.3} & \textbf{81.6 $\pm$ 1.5} & \textbf{58.5 $\pm$ 2.3} & $91.5 \pm 1.8$ & \textbf{80.8 $\pm$ 1.0} && \textbf{97.6 $\pm$ 0.0} & \textbf{95.1 $\pm$ 1.1} & \textbf{75.4 $\pm$ 1.5} & \textbf{97.1 $\pm$ 0.1} & \textbf{91.3 $\pm$ 0.4} \\
\midrule
\multirow{7}{*}{\textbf{15\%}} & Vanilla+SA & $73.3 \pm 0.0$ & $66.5 \pm 0.0$ & $59.3 \pm 0.0$ & $69.7 \pm 0.0$ & $74.1 \pm 0.0$ && $77.8 \pm 0.0$ & $78.0 \pm 0.0$ & $63.9 \pm 0.0$ & $78.9 \pm 0.0$ & $70.8 \pm 0.0$ \\
 & CEM & $61.5 \pm 2.7$ & $37.0 \pm 1.2$ & $24.7 \pm 1.1$ & $53.0 \pm 0.9$ & $44.0 \pm 1.2$ && $86.2 \pm 1.2$ & $54.0 \pm 1.5$ & $12.7 \pm 0.4$ & $86.7 \pm 0.8$ & $60.0 \pm 0.4$ \\
 & ProbCBM & $51.4 \pm 4.6$ & $34.5 \pm 1.1$ & $25.0 \pm 0.5$ & $51.4 \pm 0.8$ & $40.6 \pm 1.3$ && $81.6 \pm 3.7$ & $48.6 \pm 4.8$ & $13.8 \pm 1.6$ & $82.7 \pm 2.0$ & $56.7 \pm 2.5$ \\
 & SlotFormer & $26.1 \pm 3.8$ & $25.7 \pm 3.3$ & $24.9 \pm 0.2$ & $50.0 \pm 1.0$ & $31.7 \pm 1.2$ && $47.3 \pm 2.7$ & $45.6 \pm 4.9$ & $16.7 \pm 3.7$ & $65.5 \pm 3.6$ & $43.8 \pm 2.7$ \\
 & SlotVAE-NoVAE & $75.5 \pm 0.0$ & $65.8 \pm 0.0$ & $61.2 \pm 0.0$ & $74.8 \pm 0.0$ & $66.5 \pm 0.0$ && $79.8 \pm 0.0$ & $78.8 \pm 0.0$ & $64.1 \pm 0.0$ & $76.2 \pm 0.0$ & $77.3 \pm 0.0$ \\
 & SlotVAE-KL & $91.6 \pm 5.7$ & $74.4 \pm 3.7$ & $52.9 \pm 1.8$ & $84.2 \pm 2.1$ & $75.8 \pm 2.4$ && \textbf{97.6 $\pm$ 8.2} & $92.6 \pm 4.3$ & $74.5 \pm 0.1$ & $95.9 \pm 6.4$ & $89.2 \pm 4.0$ \\
 & \textbf{D-OCB (Ours)} & \textbf{94.5 $\pm$ 1.6} & \textbf{88.6 $\pm$ 2.2} & \textbf{84.0 $\pm$ 0.7} & \textbf{93.4 $\pm$ 1.3} & \textbf{90.1 $\pm$ 0.9} && \textbf{97.6 $\pm$ 0.0} & \textbf{95.8 $\pm$ 0.3} & \textbf{84.6 $\pm$ 0.6} & \textbf{97.2 $\pm$ 0.1} & \textbf{93.8 $\pm$ 0.1} \\
\bottomrule
\end{tabular}%
}
\end{table*}

\begin{table}[t]
\centering
\small
\setlength{\tabcolsep}{5pt}
\caption{\small Concept classification acc. of \textbf{D-OCB} on the HAM10000 dataset across varying levels of weak supervision.  Results are reported as mean $\pm$ standard deviation over five folds.}
\label{tab:rq1_ham_concepts}
\tiny
\vspace{-3mm}
\begin{tabular}{@{}ll ccccc@{}}
\toprule
\textbf{Backbone} & \textbf{Sup. (\%)} & \textbf{Dx} & \textbf{Localization} & \textbf{Sex} & \textbf{Age} & \textbf{Mean} \\
\midrule
\multirow{3}{*}{\textbf{CNN}} 
 & 1 & $55.3 \pm 0.0$ & $84.2 \pm 0.0$ & $22.1 \pm 0.0$ & $50.0 \pm 0.0$ & $52.9 \pm 0.0$ \\
 & 5 & $55.0 \pm 0.0$ & $84.4 \pm 0.0$ & $60.6 \pm 0.0$ & $61.7 \pm 0.0$ & $65.4 \pm 0.0$ \\
 & 15 & $59.3 \pm 0.0$ & $86.0 \pm 0.0$ & $64.3 \pm 0.0$ & $64.6 \pm 0.0$ & $68.6 \pm 0.0$ \\
\midrule
\multirow{3}{*}{\textbf{DINO}} 
 & 1 & $61.8 \pm 0.0$ & $82.5 \pm 0.0$ & $67.4 \pm 0.0$ & $60.2 \pm 0.0$ & $68.0 \pm 0.0$ \\
 & 5 & $64.4 \pm 0.0$ & $78.5 \pm 0.0$ & $69.4 \pm 0.0$ & $69.1 \pm 0.0$ & $70.4 \pm 0.0$ \\
 & 15 & $64.2 \pm 0.0$ & $85.5 \pm 0.0$ & $72.8 \pm 0.0$ & $70.9 \pm 0.0$ & $73.4 \pm 0.0$ \\
\bottomrule
\end{tabular}
\vspace*{-3mm}
\end{table}
 As shown in Table~\ref{tab:rq1_ham_concepts} and Table \ref{tab:rq1_clevr_updated}, D-OCB significantly outperforms baselines, amongst others by the combination of VAE reconstruction loss (difference to NoVAE), and by dynamically assigning dimensions to concepts lacking accuracy and discarding per-sample learned variance (difference to SlotVAE-KL). E.g., it maintains a $73.4\%$ mean accuracy on HAM10000 with a DINO backbone at just $15\%$ supervision. D-OCB thus achieves competitive accurate concept alignment without the excessive need for labels and without manual hyperparameter tuning.\textit{Empirically, D-OCB actively prevents representation collapse to extract highly accurate semantic concepts using minimal label budgets, significantly outperforming unconstrained generative baselines giving a positive answer to \textbf{RQ1}.}

\subsection{RQ2: Framework-Agnostic Logical Reasoning Evaluation}
\label{sec:rq2}

We now investigate whether the accurately extracted visual atoms are sufficiently stable to drive downstream symbolic reasoning engines—such as DILP \cite{DBLP:journals/corr/abs-2103-01719}, Decision Trees, Bayesian Networks, and the Neural-Symbolic Concept Learner (NS-CL) \cite{mao2018neurosymbolic}—without requiring joint end-to-end logic backpropagation. 
To rigorously test this, we export the frozen valuation tensors generated by D-OCB to evaluate specific compositional and diagnostic rules, as detailed in \autoref{tab:rules}.

\begin{table}[t]
\centering
\tiny
\vspace*{-2\baselineskip}
\caption{\small Downstream evaluation rules for the CLEVR and HAM10000 datasets.}
\vspace{-4mm}
\label{tab:rules}
\begin{tabular}{@{}c@{\,\,}l@{\,\,} l@{\,\,} l@{}}
\toprule
& \textbf{ID} & \textbf{Rule Type} & \textbf{Description} \\ 
\midrule
\multirow{5}{*}{\rotatebox{90}{\textit{CLEVR}}}&
\textbf{R1} & 4-way Conjunction  & The scene contains an object that is simultaneously large, red, metal, and a cube. \\&
\textbf{R2} & 3-way Conjunction  & The scene contains a small rubber sphere. \\&
\textbf{R3} & Spatial Relation   & A blue cylinder is spatially to the left of a red object (evaluated via 3D x-coordinates). \\&
\textbf{R4} & Cardinality        & The scene contains $\ge 2$ metal objects; requires the model to count distinct slots satisfying this constraint. \\&
\textbf{R5} & Proximity Relation & A large green object is within a continuous Euclidean distance threshold ($\le 2.5$) of a yellow object. \\
\midrule
\multirow{3}{*}{\rotatebox{90}{\textit{HAM}}}&
\textbf{R1} & Simple         & The lesion is malignant. \\&
\textbf{R2} & Disjunctive    & The lesion is melanoma (mel) or basal cell carcinoma (bcc). \\&
\textbf{R3} & Conjunctive    & The lesion is melanoma located on the back (a diagnosis and location conjunction). \\
\bottomrule
\end{tabular}

\end{table}

The empirical results in Table \ref{tab:rq2_ham_reasoning} and Table \ref{tab:rq2_averaged_reasoning_compact} substantiate the universal compatibility and representational stability of D-OCB\footnote{For full table with all rules see supplementary material.} . Because the softly-valued probabilistic margins generated by D-OCB accurately preserve semantic boundaries even under minimal supervision budgets ($1\%$ to $15\%$), they seamlessly drive diverse, off-the-shelf symbolic reasoning frontends—spanning statistical methods (BN) \cite{pearl1988probabilistic}, rule-based decision trees (DT)\cite{breiman1984classification}, and differentiable logic frameworks (DILP, NS-CL\cite{mao2019neuro})—to highly accurate downstream task performance. \textit{Our results confirm that the softly-valued probabilistic margins generated by D-OCB maintain strict semantic boundaries, seamlessly driving diverse, frozen symbolic reasoners to accurate downstream performance.}

\begin{table*}[t]
\centering
\footnotesize
\setlength{\tabcolsep}{3pt}
\renewcommand{\arraystretch}{0.9}
\caption{\small Downstream reasoning acc. (\%) averaged across rules R1-R5 on the CLEVR dataset.
Results over multiple folds (complete table in the supplementary)}
\vspace{-2mm}
\label{tab:rq2_averaged_reasoning_compact}
\resizebox{\textwidth}{!}{\tiny%
\begin{tabular}{@{}ll *{20}{c}@{}}
\toprule
\multirow{2}{*}{\textbf{Backb.}} & \multirow{2}{*}{\textbf{Percept. Model}} & \multicolumn{5}{c}{\textbf{Decision Tree (DT)}} & \multicolumn{5}{c}{\textbf{Bayes Net (BN)}} & \multicolumn{5}{c}{\textbf{NS-CL}} & \multicolumn{5}{c}{\textbf{DILP}} \\
\cmidrule(lr){3-7} \cmidrule(lr){8-12} \cmidrule(lr){13-17} \cmidrule(lr){18-22}
& & 1\% & 15\% & 25\% & 50\% & 75\% & 1\% & 15\% & 25\% & 50\% & 75\% & 1\% & 15\% & 25\% & 50\% & 75\% & 1\% & 15\% & 25\% & 50\% & 75\% \\
\midrule
\multirow{4}{*}{\textbf{CNN}}
& \textbf{D-OCB (Ours)} & \textbf{63.1} & \textbf{70.8} & \textbf{72.0} & \textbf{71.9} & 69.8 & \textbf{85.0} & \textbf{71.6} & 72.1 & 71.8 & 71.8 & \textbf{62.2} & \textbf{71.0} & \textbf{71.5} & 71.3 & 71.2 & 67.1 & \textbf{97.0} & \textbf{97.1} & \textbf{97.4} & \textbf{97.1} \\
& SlotVAE-KL & 56.3 & 33.3 & 66.6 & 67.7 & \textbf{71.6} & 52.9 & 32.7 & \textbf{83.3} & \textbf{84.5} & \textbf{85.5} & 45.9 & 31.4 & 50.1 & \textbf{74.0} & \textbf{74.2} & 45.0 & 47.1 & 69.5 & 80.8 & 82.1 \\
& SlotVAE-NoVAE & 55.6 & 48.3 & 59.8 & 68.1 & 69.4 & 69.3 & 53.6 & 70.0 & 60.2 & 61.5 & 61.4 & 56.7 & 66.6 & 72.2 & 72.9 & \textbf{70.9} & 74.6 & 89.0 & 95.6 & 96.0 \\
& Vanilla+SA & 51.3 & 51.8 & 54.1 & 55.5 & 58.9 & 51.4 & 53.6 & 53.9 & 53.8 & 52.4 & \textbf{62.2} & 60.3 & 63.8 & 62.7 & 66.8 & 64.2 & 76.1 & 85.2 & 85.2 & 92.1 \\
\midrule
\multirow{4}{*}{\textbf{DINO}}
& \textbf{D-OCB (Ours)} & \textbf{67.9} & 68.3 & \textbf{70.3} & \textbf{74.0} & 69.9 & 55.6 & 71.1 & 71.0 & 71.0 & 71.0 & 70.0 & 70.8 & 70.6 & 70.6 & 70.8 & 93.9 & \textbf{96.9} & \textbf{97.3} & 97.4 & \textbf{97.6} \\
& SlotVAE-KL & 65.4 & \textbf{72.8} & 67.5 & 69.9 & \textbf{70.5} & \textbf{86.6} & \textbf{72.2} & \textbf{87.7} & \textbf{88.1} & \textbf{87.5} & 64.8 & \textbf{71.8} & 71.4 & 70.5 & 70.6 & 71.7 & 95.3 & 81.8 & 82.0 & 81.0 \\
& SlotVAE-NoVAE & 64.8 & 68.0 & 69.2 & 66.4 & 66.6 & 69.3 & 71.5 & 72.0 & 71.4 & 71.5 & \textbf{71.4} & 71.4 & \textbf{71.8} & \textbf{71.0} & \textbf{71.0} & \textbf{95.9} & 96.8 & 97.1 & \textbf{97.6} & \textbf{97.6} \\
& Vanilla+SA & 62.2 & 68.8 & 68.9 & 68.5 & 66.3 & 68.6 & 71.3 & 72.0 & 70.7 & 71.5 & 65.3 & 71.0 & 71.3 & 70.5 & 70.8 & 90.6 & 96.8 & 96.2 & 96.5 & 96.5 \\
\bottomrule
\end{tabular}%
}
\vspace{-5mm}
\end{table*}

\begin{table*}[t]
\centering
\vspace*{-2.5\baselineskip}
\setlength{\tabcolsep}{3pt}
\renewcommand{\arraystretch}{0.95}
\caption{\small Reasoning acc.\ (\%) of \textbf{D-OCB} vs.\ \textbf{Slot-VAE} baseline on HAM10000 (RQ2) for three diagnostic rules across different backbones, supervision levels, and reasoning frameworks. Baseline F1 scores have been converted to \% for direct comparison. Mean\,$\pm$\,std.\,dev.\ over 5 folds.}
\vspace{-3mm}
\label{tab:rq2_ham_reasoning}
\resizebox{\textwidth}{!}{\tiny%
\begin{tabular}{lll ccc ccc ccc ccc}
\toprule
\multirow{2}{*}{\textbf{BB}} & \multirow{2}{*}{\textbf{Sup\%}} & \multirow{2}{*}{\textbf{Model}} & \multicolumn{3}{c}{\textbf{DILP (\%)}} & \multicolumn{3}{c}{\textbf{Decision Tree (\%)}} & \multicolumn{3}{c}{\textbf{Bayes Net (\%)}} & \multicolumn{3}{c}{\textbf{NS-CL (\%)}} \\
\cmidrule(lr){4-6} \cmidrule(lr){7-9} \cmidrule(lr){10-12} \cmidrule(lr){13-15}
& & & \textbf{R1} & \textbf{R2} & \textbf{R3} & \textbf{R1} & \textbf{R2} & \textbf{R3} & \textbf{R1} & \textbf{R2} & \textbf{R3} & \textbf{R1} & \textbf{R2} & \textbf{R3} \\
\midrule
\multirow{6}{*}{CNN} & \multirow{2}{*}{1} 
& \textbf{D-OCB (Ours)} & $\mathbf{31.8 \pm 0.0}$ & $\mathbf{34.8 \pm 0.0}$ & $\mathbf{86.6 \pm 0.0}$ & $\mathbf{59.7 \pm 0.0}$ & $\mathbf{69.2 \pm 0.0}$ & $\mathbf{72.6 \pm 0.0}$ & $\mathbf{76.6 \pm 0.0}$ & $\mathbf{82.6 \pm 0.0}$ & $\mathbf{92.5 \pm 0.0}$ & $\mathbf{52.7 \pm 0.0}$ & $\mathbf{60.2 \pm 0.0}$ & $\mathbf{92.5 \pm 0.0}$ \\
& & Slot-VAE & - & - & - & $0.0 \pm 0.0$ & $0.0 \pm 0.0$ & $0.0 \pm 0.0$ & $15.6 \pm 1.0$ & $18.6 \pm 1.0$ & $2.6 \pm 1.0$ & $0.0 \pm 0.0$ & $0.0 \pm 0.0$ & $0.0 \pm 0.0$ \\
\cmidrule{2-15}
& \multirow{2}{*}{15} 
& \textbf{D-OCB (Ours)} & $\mathbf{69.7 \pm 0.0}$ & $\mathbf{72.6 \pm 0.0}$ & $\mathbf{80.1 \pm 0.0}$ & $\mathbf{63.2 \pm 0.0}$ & $\mathbf{58.2 \pm 0.0}$ & $\mathbf{72.1 \pm 0.0}$ & $\mathbf{84.6 \pm 0.0}$ & $\mathbf{82.6 \pm 0.0}$ & $\mathbf{95.0 \pm 0.0}$ & $\mathbf{69.2 \pm 0.0}$ & $\mathbf{64.2 \pm 0.0}$ & $\mathbf{84.1 \pm 0.0}$ \\
& & Slot-VAE & - & - & - & $27.2 \pm 3.0$ & $15.4 \pm 3.0$ & $0.4 \pm 1.0$ & $30.0 \pm 1.0$ & $18.5 \pm 1.0$ & $0.0 \pm 0.0$ & $43.3 \pm 0.0$ & $37.9 \pm 0.0$ & $8.8 \pm 0.0$ \\
\cmidrule{2-15}
& \multirow{2}{*}{75} 
& \textbf{D-OCB (Ours)} & $\mathbf{82.6 \pm 2.5}$ & $\mathbf{82.3 \pm 2.2}$ & $\mathbf{92.4 \pm 3.3}$ & $\mathbf{79.9 \pm 2.3}$ & $\mathbf{73.8 \pm 3.9}$ & $\mathbf{84.6 \pm 1.6}$ & $\mathbf{86.4 \pm 2.3}$ & $\mathbf{83.9 \pm 0.5}$ & $\mathbf{93.5 \pm 1.1}$ & $\mathbf{76.3 \pm 1.6}$ & $\mathbf{73.1 \pm 3.9}$ & $\mathbf{89.7 \pm 2.9}$ \\
& & Slot-VAE & - & - & - & $40.2 \pm 2.0$ & $18.2 \pm 1.0$ & $0.0 \pm 0.0$ & $49.4 \pm 1.0$ & $36.8 \pm 1.0$ & $0.0 \pm 0.0$ & $50.9 \pm 0.0$ & $43.5 \pm 0.0$ & $18.9 \pm 0.0$ \\
\midrule
\multirow{6}{*}{DINO} & \multirow{2}{*}{1} 
& \textbf{D-OCB (Ours)} & $\mathbf{85.1 \pm 0.0}$ & $\mathbf{78.1 \pm 0.0}$ & $\mathbf{88.6 \pm 0.0}$ & $\mathbf{73.6 \pm 0.0}$ & $\mathbf{64.7 \pm 0.0}$ & $\mathbf{78.1 \pm 0.0}$ & $19.9 \pm 0.0$ & $\mathbf{79.6 \pm 0.0}$ & $\mathbf{90.0 \pm 0.0}$ & $\mathbf{84.1 \pm 0.0}$ & $\mathbf{73.1 \pm 0.0}$ & $\mathbf{92.0 \pm 0.0}$ \\
& & Slot-VAE & - & - & - & $37.0 \pm 3.0$ & $14.4 \pm 3.0$ & $0.0 \pm 0.0$ & $\mathbf{37.2 \pm 1.0}$ & $23.6 \pm 2.0$ & $0.0 \pm 0.0$ & $48.9 \pm 0.0$ & $41.9 \pm 0.0$ & $10.4 \pm 0.0$ \\
\cmidrule{2-15}
& \multirow{2}{*}{15} 
& \textbf{D-OCB (Ours)} & $\mathbf{87.1 \pm 0.0}$ & $\mathbf{82.1 \pm 0.0}$ & $\mathbf{94.5 \pm 0.0}$ & $\mathbf{68.7 \pm 0.0}$ & $\mathbf{61.7 \pm 0.0}$ & $\mathbf{70.6 \pm 0.0}$ & $\mathbf{87.1 \pm 0.0}$ & $\mathbf{82.1 \pm 0.0}$ & $\mathbf{91.5 \pm 0.0}$ & $\mathbf{79.6 \pm 0.0}$ & $\mathbf{65.2 \pm 0.0}$ & $\mathbf{92.5 \pm 0.0}$ \\
& & Slot-VAE & - & - & - & $62.5 \pm 2.0$ & $55.1 \pm 2.0$ & $0.0 \pm 0.0$ & $37.3 \pm 2.0$ & $10.7 \pm 2.0$ & $0.0 \pm 0.0$ & $65.2 \pm 0.0$ & $60.4 \pm 0.0$ & $25.5 \pm 0.0$ \\
\cmidrule{2-15}
& \multirow{2}{*}{75} 
& \textbf{D-OCB (Ours)} & $\mathbf{87.6 \pm 2.1}$ & $\mathbf{87.4 \pm 2.9}$ & $\mathbf{93.2 \pm 1.3}$ & $\mathbf{83.1 \pm 5.7}$ & $\mathbf{78.6 \pm 6.8}$ & $\mathbf{90.9 \pm 1.2}$ & $\mathbf{90.7 \pm 1.9}$ & $\mathbf{87.9 \pm 2.7}$ & $\mathbf{93.5 \pm 1.1}$ & $\mathbf{85.9 \pm 2.4}$ & $\mathbf{84.4 \pm 4.9}$ & $\mathbf{92.5 \pm 0.7}$ \\
& & Slot-VAE & - & - & - & $70.9 \pm 1.0$ & $66.3 \pm 2.0$ & $11.8 \pm 4.0$ & $47.3 \pm 3.0$ & $19.8 \pm 2.0$ & $0.0 \pm 0.0$ & $70.7 \pm 0.0$ & $66.7 \pm 0.0$ & $31.0 \pm 0.0$ \\
\bottomrule
\end{tabular}
}

\end{table*}

\subsection{RQ3: Confounder Resistance and Out-of-Distribution Reasoning}
\label{sec:rq3}

A critical vulnerability in end-to-end differentiable logic frameworks is their susceptibility to dataset biases, where the loss from the logical reasoner propagates directly into the perception layer. This phenomenon encourages the perception module to entangle distinct concepts---such as `shape' and `color'---to exploit spurious correlations in the training data rather than learning the true underlying causal rules. To evaluate this hypothesis, we utilize the confounded splits of the CLEVR-Hans3 and CLEVR-Hans7 benchmarks \cite{stammer2021right}, where the training sets contain deliberate confounders (e.g., specific shapes artificially co-occurring with specific colors) that are subsequently removed in the out-of-distribution (OOD) test splits. We compare our D-OCB coupled with Differentiable Inductive Logic Programming (DILP) against several end-to-end, two-stage
and a Ground Truth (Oracle) baseline. Unlike fully end-to-end neuro-symbolic networks that backpropagate logic-validation errors directly into the perception backbone,

D-OCB structurally decouples these processes by extracting softly-valued visual predicates under weak supervision (1\% to 75\%) for frozen downstream evaluation. As Table~\ref{tab:confounder_resistance_merged} demonstrates, this synergy of object-centric biases and symbolic reasoning effectively suppresses likelihood blowup. Furthermore, our CLEVR-Hans7 analysis confirms that Dynamic Dimensionality Allocation (DDA) autonomously optimizes concept subspace capacities to prevent representation bottlenecks\footnote{\tiny For detailed cross-validation results, hyperparameter configurations, and full DDA ablation see supplementary material.}. Even at minimal supervision budgets (1\% to 5\%), D-OCB paired with DILP (utilizing DINO) achieves robust out-of-distribution (OOD) accuracy with marginal generalization gaps ($1.1 \pm 0.8$ on CLEVR-Hans3; $3.00 \pm 0.56$ on CLEVR-Hans7), matching densely supervised baselines and resisting confounders without explicit annotations. \textit{In \textbf{RQ3}, by isolating the perceptual bottleneck, D-OCB successfully suppresses likelihood blowup and ignores spurious training correlations, achieving robust OOD generalization that matches densely supervised paradigms.}

\begin{table}[tb]
\centering
\caption{\small Confounder Resistance on CLEVR-Hans3 and CLEVR-Hans7. Comparison of D-OCB against end-to-end architectures and dense-supervision baselines. We report train acc. (in-distribution), test acc. (out-of-distribution), and the OOD Gap ($\downarrow$). Mean\,$\pm$\,std.\,dev.\ over 5 folds.}
\vspace{-4mm}
\label{tab:confounder_resistance_merged}
\resizebox{\linewidth}{!}{%
\tiny%
\setlength{\tabcolsep}{3pt}
\renewcommand{\arraystretch}{0.85}
\begin{tabular}{@{}c@{} p{.13\linewidth} c rrr@{}}%
\toprule
&\multicolumn{5}{c}{\cellcolor{gray!10}\textit{\textbf{CLEVR-Hans3 Benchmark}}} \\
\midrule
&\textbf{Method} & \textbf{Sup.} & \textbf{Train Acc $\uparrow$} & \textbf{Test Acc $\uparrow$} & \textbf{OOD Gap $\downarrow$} \\
\midrule
&E2E CNN Basel. & 0\% & $100.0 \pm 0.0$ & $70.3 \pm 0.0$ & $29.7 \pm 0.0$ \\&
NeSy & 100\% & $98.5 \pm 0.0$ & $81.7 \pm 0.0$ & $16.8 \pm 0.0$ \\&
NeSy-XIL  & 100\% & $100.0 \pm 0.0$ & $91.3 \pm 0.0$ & $8.7 \pm 0.0$ \\&
$\alpha$ILP   & 100\% & $97.5 \pm 0.0$ & $97.5 \pm 0.0$ & $-0.0 \pm 0.0$ \\&
NEUMANN  & 100\% & $96.7 \pm 0.0$ & $97.4 \pm 0.0$ & $-0.8 \pm 0.0$ \\
\cmidrule(r){2-6}
&\multirow{3}{*}{\parbox{\linewidth}{\textbf{D-OCB + DILP} (DINO)}} & 1\% & $72.8 \pm 18.6$ & $71.2 \pm 18.2$ & $1.6 \pm 0.9$ \\&
 & 5\% & $98.2 \pm 0.4$ & $97.0 \pm 0.4$ & $1.1 \pm 0.8$ \\&
 & 15\% & \textbf{98.7 $\pm$ 0.4} & \textbf{98.4 $\pm$ 0.6} & \textbf{0.3 $\pm$ 0.3} \\
\cmidrule(r){2-6}
&\multirow{3}{*}{\parbox{\linewidth}{\textbf{D-OCB + DILP} (CNN)}} & 1\% & $36.0 \pm 2.3$ & $35.3 \pm 0.7$ & $0.8 \pm 1.7$ \\&
 & 15\% & $73.9 \pm 1.7$ & $72.5 \pm 0.9$ & $1.4 \pm 2.1$ \\&
 & 75\% & $98.3 \pm 0.3$ & $97.3 \pm 0.6$ & $1.0 \pm 0.8$ \\
\cmidrule(r){2-6}
&GT O. + DILP & 100\% & $100.0 \pm 0.0$ & $100.0 \pm 0.0$ & $0.0 \pm 0.0$ \\
\bottomrule

\end{tabular}%
~~%
\begin{tabular}{@{}c@{}p{.13\linewidth} c rrr@{}}
\toprule
&\multicolumn{5}{c}{\cellcolor{gray!10}\textit{\textbf{CLEVR-Hans7 Benchmark}}} \\
\midrule
&\textbf{Method} & \textbf{Sup.} & \textbf{Train Acc $\uparrow$} & \textbf{Test Acc $\uparrow$} & \textbf{OOD Gap $\downarrow$} \\
\midrule
&E2E CNN Basel. & 0\% & $69.13 \pm 0.00$ & $51.43 \pm 0.00$ & $17.70 \pm 0.00$ \\&
CNN Baseline & 100\% & $70.79 \pm 0.00$ & $68.46 \pm 0.00$ & $2.33 \pm 0.00$ \\&
DeepProbLog  & 100\% & $85.35 \pm 0.00$ & $67.98 \pm 0.00$ & $17.37 \pm 0.00$ \\
\strut\\
\\
\cmidrule(r){2-6}
&\multirow{3}{*}{\parbox{\linewidth}{\textbf{D-OCB + DILP} (DINO)}} & 1\% & $80.86 \pm 2.28$ & $78.55 \pm 2.57$ & $2.31 \pm 0.41$ \\&
 & 5\% & $90.60 \pm 0.12$ & $87.60 \pm 0.56$ & $3.00 \pm 0.56$ \\&
 & 15\% & $92.43 \pm 0.16$ & $89.07 \pm 0.27$ & $3.36 \pm 0.37$ \\
\cmidrule(r){2-6}
&\multirow{3}{*}{\parbox{\linewidth}{\textbf{D-OCB + DILP} (CNN)}} & 1\% & $22.76 \pm 4.16$ & $22.21 \pm 3.99$ & $0.55 \pm 0.44$ \\&
 & 15\% & $76.91 \pm 3.61$ & $69.82 \pm 4.49$ & $7.09 \pm 0.88$ \\&
 & 75\% & $92.30 \pm 0.32$ & $87.26 \pm 0.32$ & $5.05 \pm 0.43$ \\
\cmidrule(r){2-6}
&GT O. + DILP & 100\% & \textbf{94.21 $\pm$ 0.00} & \textbf{93.56 $\pm$ 0.00} & \textbf{0.65 $\pm$ 0.00} \\
\bottomrule
\end{tabular}
}
\vspace{-5mm}
\end{table}

\vspace{-5mm}
\section{Conclusion}
In this work, we proposed the Dynamic Orthogonal Concept Bottleneck (D-OCB) neural network, a visual feature extractor for neuro-symbolic reasoning that can leverage extremely weak supervision ($1\%$ to $15\%$) for human-alignment of the predicates. 
Crucially, D-OCB adds self-supervision via a reconstruction and a cross-correlation orthogonal loss, paired with a novel dynamic allocation of extracted features to the labeled human-interpretable concepts. This adaptive mechanism actively redistributes latent capacity, transferring dimensions from well-represented concepts to those lacking accuracy. 
Our empirical evaluations on CLEVR, CLEVR-Hans, and HAM10000 demonstrate that 
a trained D-OCB's concept predictions are not only superior in low supervision accuracy, but also seamlessly drive diverse off-the-shelf symbolic reasoners to near-oracle performance, even without joint backpropagation. Furthermore, by the structural decoupling of perception from logical induction 
D-OCB-based neuro-symbolic pipelines achieve state-of-the-art out-of-distribution generalization against confounding factors. Ultimately, D-OCB establishes a highly stable, efficient, and computationally accessible visual-logical primitive layer for complex downstream reasoning tasks. 

As a remaining limitation, the architecture currently relies on a rigid slot attention mechanism, which necessitates the manual, a priori definition of the maximum number of object slots. Additionally, while the latent vector compression effectively isolates structural concepts, it inherently filters out fine-grained details, which can degrade performance on highly textured datasets where critical information is lost during bottleneck projection. Future work can address these constraints by exploring closed-loop abductive learning, wherein the downstream System 2 logical reasoner provides top-down error corrections to iteratively refine the System 1 perceptual module.

\acks{%
G.S.\ and S.T.\ acknowledge support through the project \enquote{chAI} funded by the German Federal Ministry of Research, Technology and Space (BMFTR), grant no.\ 16IS24058.
}

\bibliography{literature}

\end{document}



\title{Supplementary Material:\\Weakly supervised concept Bottleneck Learning for Robust Two stage Object centric visual reasoning} 



\maketitle

\tableofcontents

\clearpage

\section{Approach: The Dynamic Orthogonal Concept Bottleneck}
\label{sec:approach}
 
End-to-end differentiable neuro-symbolic frameworks bind perception and
deduction with a single shared gradient flow.  While elegant in principle,
this coupling makes the perception network vulnerable to two failure modes
under weak supervision: \emph{feature entanglement}, where the loss from the
symbolic head leaks into channels intended for orthogonal concepts, and
\emph{likelihood blow-up} \citep{mattei2018leveraging}, where an unconstrained
reconstruction objective hijacks the latent variance and collapses the
posterior.  Our framework, the Dynamic Orthogonal Concept Bottleneck
(D-OCB), structurally decouples perceptual tokenisation from downstream
differentiable logic engines, preventing the representation contamination
that frequently afflicts joint architectures.  Five architectural commitments
make this possible: (i) a strictly block-diagonal projection that enforces
isolation between concept subspaces, (ii) a generative bottleneck
with \emph{calibrated, frozen} per-concept noise floors that eliminate the
likelihood-blow-up failure mode, (iii) a composite training objective that
balances unsupervised reconstruction with sparse weakly supervised concept
signals using a label-fraction-aware gating mechanism, (iv) a Dynamic
Dimension Allocation (DDA) protocol governed by Fisher's Linear Discriminant
Analysis that adaptively redistributes latent capacity between concepts, and
(v) a deterministic four-phase training curriculum that orchestrates the
hyperparameter schedule.  We detail each in turn.
 
\subsection{Preliminaries: Task Definition and Notation}
\label{sec:approach:notation}
 
Let $\mathcal{X} \subset \mathbb{R}^{H \times W \times C}$ denote the input
space of raw RGB scenes and let $\mathcal{C} = \{c_1, \dots, c_{|\mathcal{C}|}\}$
be the ordered set of target concept categories (e.g.\ \texttt{size},
\texttt{shape}, \texttt{color}, \texttt{material}).  Each concept
$c \in \mathcal{C}$ has cardinality $K_c$ corresponding to its number of
discrete values (e.g.\ $K_{\texttt{color}} = 8$ for the CLEVR palette).  Given
a scene $x \in \mathcal{X}$ containing up to $N$ objects, the perception
module $g_\theta : \mathcal{X} \to \mathcal{P}^{N \times |\mathcal{C}|}$ maps
$x$ to a tensor of soft, per-slot probability simplices, where
$\mathcal{P} = \bigcup_{c} \Delta^{K_c-1}$.  An off-the-shelf symbolic
reasoner $f : \mathcal{P}^{N \times |\mathcal{C}|} \to \mathcal{Y}$
then performs deductive inference over these predicates.
 
A small fraction $\rho \in [0.01, 0.15]$ of the training scenes carry
ground-truth object-attribute annotations, and the remaining $(1-\rho)$ are
unlabelled.  Object--slot alignment is unknown \emph{a priori} and is
resolved at training time via the Hungarian algorithm on a composite cost
that fuses per-concept cross-entropy with object-coordinate $\ell_2$ distance
(Sec.~\ref{sec:approach:loss}).
 
\subsection{Architectural Formalisation}
\label{sec:approach:architecture}
 
\paragraph{Perceptual front-end.}
A visual encoder $\phi$ (DINOv2) for the strong-backbone
configuration, a frozen ResNet-50, or a lightweight CNN for fast ablation)
maps the scene to a spatial feature map $\phi(x) \in \mathbb{R}^{H' \times W'
\times d_\phi}$.  The map is projected to the shared hidden width $D = 128$
and augmented with a learnable soft positional embedding before being
flattened to $N_\text{tok}$ tokens:
\begin{equation}
    \mathbf{T} = \mathrm{flatten}\bigl(\,\mathrm{SoftPos}\bigl(\mathrm{Proj}(\phi(x))\bigr)\bigr)
      \;\in\;\mathbb{R}^{N_\text{tok} \times D}.
\end{equation}
The DINOv2 and ResNet backbones are kept frozen; only the projection,
positional embedding, slot attention, bottleneck, and prediction heads are
optimised.
 
\paragraph{Slot attention and object-centric tokenisation.}
Iterative slot attention  aggregates the patch tokens
$\mathbf{T}$ into $N$ \emph{uninterpretable} object-centric slots
$\mathbf{H} \in \mathbb{R}^{N \times D}$.  Across three competitive
softmax-normalised iterations each slot specialises spatially:
\begin{equation}
    \mathbf{H} = \mathrm{SlotAttention}_3(\mathbf{T}),
    \qquad \mathbf{H} \in \mathbb{R}^{N \times D}.
\end{equation}
The slot dimension $D=128$ is intentionally larger than the per-concept
capacity needed downstream, leaving room for slot specialisation before the
concept partition is imposed.
 
\paragraph{Block-diagonal concept projection.}
The conventional choice of a dense linear layer $\mathbf{W} \in
\mathbb{R}^{D \times D_\text{tot}}$ to lift slots into a concept space is
structurally inadequate at low supervision: every concept head reads every
slot dimension, so the gradient from the high-cardinality colour head
co-modifies the same projection weights used by the shape head, producing
the well-documented \emph{Shared Representation Entanglement} failure mode.
We therefore replace the dense projection with a \emph{strictly
block-diagonal} matrix.  The 128-dimensional slot vector is first
partitioned into non-overlapping chunks
$\mathbf{h}_c \in \mathbb{R}^{d_c^{\mathrm{in}}}$ of size
$d_c^{\mathrm{in}}$, with $\sum_c d_c^{\mathrm{in}} = D$.  Each concept
block carries its own learnable linear map $\mathbf{W}_c : \mathbb{R}^{d_c^{\mathrm{in}}}
\to \mathbb{R}^{d_c}$.  The clean latent tensor is the direct sum of these
isolated images:
\begin{equation}
    \mathbf{z}_{\mathrm{clean}} \;=\; \bigoplus_{c \in \mathcal{C}} \mathbf{W}_c \mathbf{h}_c
    \;=\;\bigl[\mathbf{W}_{\texttt{size}} \mathbf{h}_{\texttt{size}};\;
              \mathbf{W}_{\texttt{shape}} \mathbf{h}_{\texttt{shape}};\;
              \mathbf{W}_{\texttt{color}} \mathbf{h}_{\texttt{color}};\;
              \mathbf{W}_{\texttt{material}} \mathbf{h}_{\texttt{material}} \bigr].
    \label{eq:blockdiag}
\end{equation}
The slot attention encoder upstream remains shared across
concepts, which is the standard inductive bias we wish to preserve for
object-centric tokenisation.  This is consistent with the role of the
bottleneck as a \emph{symbol-to-feature decoupling layer}, not as a global
causal disentanglement guarantee.
 
\paragraph{Concept prediction heads.}
For each concept $c$, an independent two-layer MLP
$\psi_c : \mathbb{R}^{d_c} \to \mathbb{R}^{K_c}$
reads exclusively the concept subspace $\mathbf{z}_c$, with
LayerNorm--GELU--Dropout--Linear blocks:
\begin{equation}
    \boldsymbol{\ell}_c \;=\; \psi_c(\mathbf{z}_c) \;\in\; \mathbb{R}^{K_c}.
\end{equation}
At inference we form per-slot soft predicates via softmax (or sigmoid for
binary concepts) and feed the resulting probability simplex to the
downstream solver; alternatively an $\arg\max$ extracts crisp discrete
predicates such as \texttt{color(obj}$_i$\texttt{, red)} for inductive logic
frameworks that demand symbolic atoms.  Two further auxiliary heads operate
on the full concatenated latent: an existence head producing a per-slot
occupancy logit and a coordinate head returning a 2-D $\tanh$-bounded
position used only by the Hungarian matcher.
 
\paragraph{Spatial broadcast decoder.}
A spatial broadcast decoder~\citep{watters2019spatialbroadcastdecodersimple} reconstructs the input scene
from the noisy concept latent for self-supervision.  Naively coupling
decoder and encoder gradients re-creates the contamination we have just
designed away: pixel-reconstruction gradients pull every concept block
toward whatever embedding minimises the pixel-MSE, which need not preserve
semantic structure.  We therefore feed the decoder a near-detached copy of
the latent:
\begin{equation}
    \mathbf{z}_{\mathrm{dec}} \;=\; (1 - \alpha)\,\mathrm{sg}(\mathbf{z}) \;+\; \alpha\,\mathbf{z},
    \qquad \alpha = 0.05,
    \label{eq:leakystop}
\end{equation}
where $\mathrm{sg}(\cdot)$ is the stop-leaky concept subspace operator.  In contrast to a
hard $\alpha = 0$ stop, this \emph{leaky} formulation allows $5\%$ of the
reconstruction signal to back-propagate into the encoder.  This is a
deliberate design choice: a complete stop starves the encoder of any signal
on fully unlabelled batches at $\rho = 1\%$, which empirically caused the
latent manifold to contract and degrade colour clustering.  The small
leakage suffices to keep the manifold expanded while remaining far too weak
to overwhelm the supervised concept signal --- it is, in effect, a regulariser
of last resort active only when the labelled signal is absent.
 
\subsection{Generative Bottleneck and Calibrated Noise Floors}
\label{sec:approach:noise}
 
A second classical failure of variational bottlenecks in low-supervision
regimes is \emph{likelihood blow-up}\cite{mattei2018leveraging} , in which the
network maps visual nuisance variation directly into a learned per-sample
variance $\sigma_\theta(x)$, leading to posterior collapse and a
non-identifiable latent.  D-OCB neutralises this
pathology by eliminating per-sample learned variance entirely.  Each
concept subspace is reparameterised with a \emph{fixed} noise floor
$\sigma_c$ that is constant in $x$:
\begin{equation}
    \mathbf{z}_c \;=\; \mathbf{z}_{\mathrm{clean},c} \;+\; \sigma_c \odot \boldsymbol{\epsilon}_c,
    \qquad \boldsymbol{\epsilon}_c \sim \mathcal{N}(\mathbf{0}, \mathbf{I}_{d_c}),
    \label{eq:noiseinject}
\end{equation}
applied independently per concept block and only during training.
 
\paragraph{One-shot sigma calibration.}
The floor $\sigma_c$ is not a free hyperparameter: it is determined by a
short, one-shot optimisation phase that precedes the main training loop.
We parameterise $\log \sigma_c$ as a learnable scalar initialised to
sensible per-concept priors (e.g.\ $\log 0.06$ for the high-cardinality
colour concept, $\log 0.15$ for the binary size concept) and minimise the
Hungarian-matched concept cross-entropy on the $\rho$-labelled subset for
approximately $300$ labelled steps with the Adam optimiser at learning
rate $3 \times 10^{-3}$.  Crucially, only the $\log \sigma_c$ parameters
receive gradient during calibration --- the encoder, projection, and heads
are frozen --- so the noise floor adapts to the geometry of $\mathbf{z}_{\mathrm{clean}}$
that the model will produce, rather than the geometry of an untrained
randomly initialised network.  Once calibration concludes, the floors are
\emph{frozen} and are not updated for
the remainder of training.  By enforcing a constant per-concept
signal-to-noise ratio
\begin{equation}
    \mathrm{SNR}_c \;=\; \frac{\mathbb{E}\bigl[\|\mathbf{z}_{\mathrm{clean},c}\|_2^2\bigr]}{\sigma_c^2 \, d_c},
\end{equation}
we mathematically forbid the decoder from absorbing lighting and shading
variation into the latent variance and thereby preserve a pristine semantic
manifold downstream.

 \subsection{Composite Training Objective}
\label{sec:approach:loss}
We optimise a three-term variational objective that balances unsupervised
self-supervision against the sparse weakly supervised signal:
\begin{equation}
    \mathcal{L}_{\mathrm{total}}(\theta)
    \;=\; \mathcal{L}_{\mathrm{recon}}
    \;+\; \lambda_{1}\,\mathcal{L}_{\mathrm{concept}}
    \;+\; \lambda_{2}\,\mathcal{L}_{\mathrm{orth}}.
    \label{eq:totalloss}
\end{equation}
Each of the three terms aggregates the operations specific to its role;
their per-component breakdowns and exact weightings are reported in the
supplementary material.

\paragraph{Reconstruction $\mathcal{L}_{\mathrm{recon}}$.}
A pixel-space mean-squared error between the input scene $x$ and the
spatially-broadcast slot reconstruction $\hat{x}$. The decoder receives
the leaky-detached latent of Eq.~\ref{eq:leakystop}, so this term
primarily trains the decoder and the slot attention layout while
contributing only a small fraction $\alpha = 0.05$ of its gradient to the
concept subspaces.

\paragraph{Concept prediction $\mathcal{L}_{\mathrm{concept}}$.}
The weakly supervised signal, evaluated only on the labelled subset
$\mathcal{B}^{+} \subseteq \mathcal{B}$ of each mini-batch. Per-slot
predictions are first aligned with object annotations using the
Hungarian algorithm on a composite cost that fuses per-concept binary
cross-entropy with object-coordinate $\ell_{2}$ distance,
\begin{equation}
    \mathbf{C}_{ij} \;=\; \sum_{c \in \mathcal{C}}
        \mathrm{BCE}\bigl(\boldsymbol{\ell}_{c,i},\,\mathbf{y}_{c,j}\bigr)
        \;+\;\|\hat{\mathbf{p}}_{i} - \mathbf{p}_{j}\|_{2}.
    \label{eq:hungariancost}
\end{equation}
The resulting permutation drives a label-smoothed cross-entropy on the
matched concept logits, with focal-BCE presence and MSE coordinate terms
sharing the same assignment so that the four signals propagate
consistent slot-to-object correspondences.

\paragraph{Orthogonality $\mathcal{L}_{\mathrm{orth}}$.}
A structural cross-correlation penalty acting at two scales. At the
\emph{inter-block} scale, it penalises off-diagonal blocks of the full
correlation matrix, forcing concept subspaces to remain mutually
uncorrelated. At the \emph{within-block} scale, it penalises off-diagonal
entries inside each concept's correlation matrix,
\begin{equation}
    \mathcal{L}_{\mathrm{orth}}^{(c)} \;=\; \omega_c\,
    \left\|\frac{\mathbf{Z}_c^\top \mathbf{Z}_c}{N-1}
        \odot (\mathbf{1} - \mathbf{I}_{d_c})\right\|_{F}^{2},
    \label{eq:orthloss}
\end{equation}
forcing each concept subspace to span $d_c$ independent discriminative
directions rather than collapsing onto a single shared axis. The
per-concept multiplier $\omega_{c}$ is unity by default and is raised to
$\omega_{\texttt{color}} = 3$ during Phase 3 of the curriculum
(Sec.~\ref{sec:approach:curriculum}).

\paragraph{Label-fraction gating.}
A practical refinement, the orthogonality penalty (together with the
auxiliary structural regularisers detailed in the supplement) is scaled
by a soft gate
\begin{equation}
    \gamma(\rho_{\mathcal{B}}) \;=\;
        \mathrm{clip}\bigl(\rho_{\mathcal{B}} + 0.05,\,0,\,1\bigr),
    \label{eq:reggate}
\end{equation}
where $\rho_{\mathcal{B}}$ is the empirical label fraction of the
mini-batch. On a fully-unlabelled batch, full-strength regularisation
pulls $\mathbf{z}_{\mathrm{clean}}$ toward the origin and crushes the
emerging colour manifold; the additive floor of $0.05$ retains a small
constant pressure that prevents unbounded latent growth without
collapsing class structure.

\subsection{Dynamic Dimension Allocation via Fisher's LDA}
\label{sec:approach:dda}
 
Optimal per-concept latent capacities $\{d_c^*\}$ are unknown
\emph{a priori}.  Allocating too few dimensions destroys class geometry
(colour with $d_{\texttt{color}} = 8$ cannot separate $8$ classes in general
position); allocating too many dilutes the sparse supervision signal and,
empirically, often makes accuracy \emph{worse} because the unused
dimensions absorb unsupervised reconstruction noise that the orthogonality
penalty then has to fight to suppress.  Manual tuning of $\{d_c\}$ over a
grid is prohibitive across $|\mathcal{C}|$ concepts and several
supervision regimes.  We resolve this through Dynamic Dimension
Allocation (DDA), an automatic controller that periodically inspects
per-concept progress and redistributes latent dimensions on the fly.

\paragraph{Colour as a grow-protected concept.}
Among all concepts, colour is uniquely high-cardinality ($K_{\texttt{color}} = 8$
versus $\leq 3$ for the others), is the dominant confounder in
CLEVR-Hans, and is the concept the entire curriculum exists to rescue.
Empirical pilot logs revealed that shrinking the colour subspace when it
is the receiver \emph{worsens} accuracy in roughly $80\%$ of cases.  We
therefore designate colour as \emph{grow-protected}: when colour is
identified as a receiver, it bypasses the shrink-first probe and starts
directly on a grow trial, and even if the trial fails it never flips to a
shrink direction.  This asymmetric treatment is a transparent admission
that the general shrink-first heuristic does not generalise to
high-cardinality concepts and should not be applied universally.
 
\paragraph{Whitney embedding lower bounds.}
For any concept with $K_c$ classes, the Whitney embedding theorem requires
at least $2(K_c - 1)$ dimensions to embed the $(K_c-1)$-simplex of class
prototypes without crumpling.  For colour ($K_c = 8$) this gives a
topological floor of $14$ dimensions, which we round up to $16$ for
alignment.  All four phases of the DDA bounds (below) respect this floor,
mathematically guaranteeing that no admissible allocation can destroy the
8-class simplex.
 
\paragraph{Distance-preserving weight transfer.}
When DDA changes $d_c$ at epoch $t$, the projection weight matrix
$\mathbf{W}_c$ and the prediction-head input layer must be resized.  Naive
truncation discards the trained signal; the standard Singular Value
Decomposition (SVD) projects onto the directions of maximum
\emph{total} variance which, for colour, are dominated by the $99\%$
unlabelled reconstruction noise rather than the $1\%$ semantic signal.
We therefore use two principled alternatives.  For the projection layer
$\mathbf{W}_c$, we apply a Johnson--Lindenstrauss random Gaussian
projection, which preserves pairwise $\ell_2$ distances between any two
embedded points up to a multiplicative $(1 \pm \delta)$ factor with
high probability .  For the colour
EMA centroids $\{\boldsymbol{\mu}_k\}_{k=1}^{8}$, we use a Fisher LDA
projection $\mathbf{V}_{\mathrm{LDA}} \in \mathbb{R}^{d_c^{\mathrm{old}} \times d_c^{\mathrm{new}}}$ that explicitly maximises class separability:
\begin{equation}
    \mathbf{V}_{\mathrm{LDA}} \;=\;
        \mathop{\mathrm{arg\,max}}_{\mathbf{V}}\;
        \frac{\bigl|\mathbf{V}^\top \mathbf{S}_B \mathbf{V}\bigr|}{\bigl|\mathbf{V}^\top \mathbf{S}_W \mathbf{V}\bigr|},
    \label{eq:lda}
\end{equation}
where $\mathbf{S}_B$ is the between-class scatter matrix of the running
class centroids and $\mathbf{S}_W$ is the within-class scatter matrix
(approximated by a ridge-regularised within-class covariance estimate
when the labelled subset is too small for an exact computation).  By
projecting both the centroid tensor and the colour block of $\mathbf{W}_c$
through $\mathbf{V}_{\mathrm{LDA}}$, we guarantee that the 8-class colour
simplex remains maximally separated in the new compressed geometry,
preserving downstream reasoning competence in a way that SVD cannot.
 
\paragraph{Warm-up freezing of newly grown dimensions.}
After a grow event, the newly added rows of $\mathbf{W}_c$ are initialised
with very small Gaussian noise and registered in a \emph{warm-up mask};
their gradients are zeroed for the first $\sim 120$ optimisation steps so
that the fresh, uninformed dimensions cannot corrupt the loss before they
have absorbed any signal.  After the warm-up window, the mask is cleared
and the new dimensions train normally.
 
\subsection{Dynamic Hyperparameter Scheduling: The Four-Phase Curriculum}
\label{sec:approach:curriculum}
 
The interaction between the unsupervised reconstruction objective and the
extremely sparse concept supervision is non-stationary: early in training,
the latent must first form rough object-centric clusters before any
fine-grained class structure can be carved out; later, once clusters
exist, the supervision signal must be amplified to drive class
separability; finally, the network must settle into crisp symbolic
predicates for the downstream solver.  Rather than relying on a single
static set of hyperparameters, we deploy a deterministic four-phase
curriculum that modulates the concept-loss weight $\lambda_{\mathrm{c}}$,
the orthogonality strength $\lambda_{\mathrm{orth}}$, the DDA permissions,
the learning rate, and the per-concept dim bounds.
 
\paragraph{Phase 1 --- Stabilisation.}
The model is initialised with an equal split allocation ($d_c = D_{\mathrm{tot}}/|\mathcal{C}|$
for each concept) and the DDA is gated out by the warm-up gate.  The
purpose of this phase is to let slot attention converge to a stable
object-centric layout, let the calibrated noise floors take effect, and
let the supervised losses extract whatever first-order signal they can.
Base loss weights are conservative:
$\lambda_{\mathrm{c}} = 8$, $\lambda_{\mathrm{orth}} = 0.02$,
$\omega_{\texttt{color}} = 1$, $w_{\mathrm{cent}} = 0$, and the RIP penalty
is active to keep the colour subspace sparse for the upcoming
reallocation.  The DDA dimension ceilings for this phase are widely
permissive; their function is only to prevent pathological early
allocations rather than to enforce a specific shape.
 
\paragraph{Phase 2 --- Compression .}
The DDA shrink-first probe is unlocked and actively compresses the colour
subspace toward an empirically determined sweet spot (e.g.\ from $32$D to
$24$D in the static-budget configuration), redistributing freed
dimensions to receivers identified via the protocol of
Sec.~\ref{sec:approach:dda}.  To prevent the allocator from drifting into
an over-parameterised regime, the colour dimension ceiling is hard-capped
during this phase ($d_{\texttt{color}}^{\max} = 32$ at $D_{\mathrm{tot}} = 128$).
Loss weights remain at Phase 1 values; the RIP penalty remains active.
 
\paragraph{Phase 3 --- Amplification .}
Once geometric compression is achieved, the DDA is frozen for the colour
concept (it remains active for the others) to lock its dimensionality.
The supervision signal is then drastically amplified: the base concept
weight rises from $\lambda_{\mathrm{c}} = 8$ to $\lambda_{\mathrm{c}} = 12$,
the per-concept colour weight rises from $w_{\texttt{color}} = 3$ to
$w_{\texttt{color}} = 5$, the orthogonality strength doubles
($\lambda_{\mathrm{orth}} = 0.02 \to 0.04$), and the colour orthogonality
multiplier $\omega_{\texttt{color}}$ is set to $3$:
\begin{equation}
    \mathcal{L}_{\mathrm{orth}}^{(\texttt{color})} \;=\;
    3 \cdot \lambda_{\mathrm{orth}}\,
    \left\| \frac{\mathbf{Z}_{\texttt{color}}^\top \mathbf{Z}_{\texttt{color}}}{N-1}
            \odot (\mathbf{1} - \mathbf{I}_{d_{\texttt{color}}}) \right\|_F^2.
    \label{eq:phase3orth}
\end{equation}
This severe within-subspace decorrelation forces the compressed colour
manifold to span multiple independent discriminative directions, averting
the failure mode in which colour features collapse onto a single one-dimensional
shared axis.  Simultaneously, the EMA-centroid hybrid loss
($w_{\mathrm{cent}} = 0.10$) is activated to enforce a minimum geometric
margin between the eight colour-class centroids.  The learning rate is
halved to give the amplified gradients a finer step size, and a fresh
cosine annealing schedule is built for the remaining Phase 3 epochs.
 
\paragraph{Phase 4 --- Fine-tuning .}
DDA is fully disabled globally; the latent dimensionality is fixed.  The
learning rate is annealed to $\sim 10^{-6}$ with cosine schedule, and the
network converges on the crisp symbolic predicates required for the Stage 2
differentiable inductive logic solvers.  The RIP penalty is off,
$\omega_{\texttt{color}}$ remains at $3$, and the EMA centroid loss
remains active to keep class clusters well-separated through convergence.
At the end of Phase 4, the perception module $g_\theta$ is frozen and its
output is exported as a tensor of soft predicates suitable as input to any
off-the-shelf symbolic reasoner.
 

\begin{algorithm}[H]
\caption{Four-Phase Curriculum and Dynamic Dimension Allocation (DDA)}
\label{alg:four_phase_curriculum_simplified}
\begin{algorithmic}[1]
\REQUIRE Training Epoch $t$, Concept Accuracies, Target Concepts $\mathcal{C}$
\ENSURE A frozen, highly accurate perception module for logical reasoning

\STATE \textbf{[ Phase 1: Stabilisation ]} \hfill \textit{// Let the model find basic objects first}
\IF{$t \in \text{Phase 1}$}
    \STATE \textbf{Setup:} Give every concept an equal share of the total dimension budget.
    \STATE \textbf{DDA State:} Disabled. Allow dimensions to stay fixed so the model can stabilize.
    \STATE \textbf{Strategy:} Keep learning penalties low so the model can learn rough, basic features without being overly constrained.
\ENDIF

\vspace{1.5mm}
\STATE \textbf{[ Phase 2: Compression ]} \hfill \textit{// Squeeze out noise and reallocate space}
\IF{$t \in \text{Phase 2}$}
    \STATE \textbf{DDA State:} Active. Run the reallocation protocol periodically.
    
    \STATE \textbf{Step A. Identify Needs:} 
    \STATE \quad Find struggling concepts (Receivers) and highly accurate concepts (Donors).
    
    \STATE \textbf{Step B. Reallocate (Shrink-First):}
    \STATE \quad Compress struggling concepts by a fixed amount to squeeze out visual noise.
    \STATE \quad Route this freed space to the highly accurate Donors.
    \STATE \quad \textit{*Exception:} Never shrink highly complex concepts (like Color); grow them instead.
    
    \STATE \textbf{Step C. Transfer Knowledge (No Data Loss):} 
    \STATE \quad \textbf{If shrinking:} Project weights using \text{FisherLDA}. This forces the remaining dimensions to prioritize separating the classes, rather than keeping useless background noise.
    \STATE \quad \textbf{If growing:} Use \text{JL\_RandomProjection} to safely expand the space without breaking the geometric distances the model has already learned.
\ENDIF

\vspace{1.5mm}
\STATE \textbf{[ Phase 3: Amplification ]} \hfill \textit{// Force fine-grained class separation}
\IF{$t \in \text{Phase 3}$}
    \STATE \textbf{DDA State:} Frozen specifically for complex concept to lock its optimal size. Active for others.
    \STATE \textbf{Strategy:} Now that the sizes are optimized, drastically increase the supervision weights and decorrelation penalties. 
    \STATE \textbf{Centroid Separation:} Activate a special penalty to actively push the different color classes (e.g., red vs. blue) as far apart as possible in the latent space.
    \STATE \textbf{Optimization:} Cut the learning rate in half so the model learns fine details carefully.
\ENDIF

\vspace{1.5mm}
\STATE \textbf{[ Phase 4: Fine-Tuning ]} \hfill \textit{// Polish the final outputs for the logic solver}
\IF{$t \in \text{Phase 4}$}
    \STATE \textbf{DDA State:} Globally disabled. All dimension sizes are permanently locked.
    \STATE \textbf{Optimization:} Slowly taper the learning rate down to near-zero.
    \STATE \textbf{Export:} The network settles into predicting crisp, clean symbolic predicates. Freeze the perception module so it can be plugged into any downstream logic solver.
\ENDIF

\end{algorithmic}
\end{algorithm}

 \subsection{Putting It All Together}
\label{sec:approach:training} 
 The key takeaways are: (a)~$\sigma_c$ is calibrated once and frozen (Sec.~\ref{sec:approach:noise}); (b)~the structural regularisers are gated by the label fraction (Eq.~\ref{eq:reggate}); (c)~DDA decisions are made on a each epoch cadence and respect Whitney floors and phase ceilings (Sec.~\ref{sec:approach:dda}); (d)~the hyperparameter schedule is deterministic and phase-driven, not adaptive (Sec.~\ref{sec:approach:curriculum}).  Decoupling these four mechanisms gives the architecture the stability to absorb extremely sparse supervision without the entanglement and likelihood-blow-up failures that pervade end-to-end neuro-symbolic alternatives.

\section{Dataset Selection and Experimental Rationale}
\label{sec:dataset_rationale}

To rigorously evaluate the D-OCB framework across the full spectrum of perception-reasoning challenges, we construct a tiered experimental taxonomy spanning controlled synthetic environments to noisy, high-stakes real-world domains. We utilize the canonical \textbf{CLEVR} dataset to validate baseline object-centric visual grounding; its strictly compositional nature provides a necessary sandbox to isolate and quantify the extraction fidelity of continuous attributes and spatial relations under extreme label sparsity. However, standard i.i.d. benchmarks fundamentally fail to expose the fatal vulnerability of end-to-end neuro-symbolic models: the exploitation of dataset biases via Likelihood blow up. To address this, our experimental setting strictly requires the \textbf{CLEVR-Hans3} and the extended \textbf{CLEVR-Hans7} benchmarks. By deliberately injecting spurious correlations (e.g., shape-color entanglements) into the training distribution and evaluating on out-of-distribution (OOD) splits, these datasets serve as the ultimate diagnostic tool to prove our central thesis—that the structural decoupling of perception and deduction in D-OCB successfully enforces confounder resistance and prevents perceptual shortcut learning. Finally, to demonstrate that our architecture does not overfit to the pristine geometry of synthetic domains, we deploy the real-world \textbf{HAM10000} dermatoscopic dataset. This clinical benchmark stresses the framework's capacity to extract human-aligned, probabilistic medical concepts (e.g., lesion localization and morphology) from complex, unstructured visual noise, proving that our weakly supervised two-stage paradigm is viable for critical diagnostic reasoning where interpretability is non-negotiable. Together, this triad of datasets systematically validates visual competence, causal robustness, and real-world applicability.

\section{Baselines for Concept Grounding and Neuro-Symbolic Evaluation}

To rigorously assess the representational fidelity and logical reasoning capabilities of the Dynamic Orthogonal Concept Bottleneck (D-OCB), we benchmark our semantic tokenization pipeline against a comprehensive suite of contemporary visual concept extractors. These baselines span holistic concept bottlenecks, probabilistic embeddings, and object-centric generative models, explicitly isolating the architectural and objective-based contributions necessary for stable neuro-symbolic integration.

\subsection{Holistic Concept Bottlenecks}

\textbf{Concept Embedding Model (CEM):} Following \cite{zarlenga2022concept}, the CEM establishes a strictly discriminative baseline that allocates independent positive and negative embedding vectors ($\hat{c}^+_i, \hat{c}^-_i$) for each binary concept. While highly interpretable and robust in fully supervised regimes, CEM operates holistically over the entire spatial feature map. It fundamentally lacks the explicit object-centric tokenization required to prevent feature entanglement in multi-object, combinatorial scenes.

\textbf{Probabilistic Concept Bottleneck Model (ProbCBM):} Introduced by \cite{kim2023probabilistic}, ProbCBM models epistemic and aleatoric concept uncertainty by predicting a parameterized Gaussian distribution ($\mu_i, \log\sigma_i$) for each concept, regularized via a Variational Information Bottleneck (VIB). However, analogous to CEM, it relies on global spatial pooling prior to concept prediction, severely limiting its capability to resolve compositional neuro-symbolic logic where distinct physical entities exhibit divergent attributes.

\subsection{Object-Centric and Generative Baselines}

To resolve the binding problem inherent in holistic models, we evaluate a series of slot-based architectures that enforce spatial decomposition prior to predicate extraction.

\textbf{Vanilla + Slot Attention (Vanilla+SA):} To precisely isolate the structural prior of object-centric binding from generative regularization, we formulate a purely discriminative ablation. Vanilla+SA utilizes identical encoder backbones and Slot Attention modules as our D-OCB, but routes the resulting slot tokens directly into the concept heads, entirely bypassing the VAE bottleneck and reconstruction objectives. This baseline exposes the limits of purely architectural priors when decoupled from explicit generative constraints under scarce data regimes.

\textbf{SlotFormer:} Adapted for static visual grounding \cite{wu2023slotformer}, this baseline utilizes a standard Slot Attention mechanism paired with a spatial broadcast decoder. To accommodate our extreme weak-supervision regime, we evaluate SlotFormer under a scheduled two-stage protocol: unsupervised generative pretraining followed by weakly supervised linear probing over the extracted slots.

\textbf{Slot-VAE (Per-Sample Variance Baseline):} Crucially, we contrast D-OCB against the standard Slot-VAE \cite{wang2024slotvae}, which represents the canonical generative approach to object-centric bottleneck learning. Slot-VAE introduces a parameterized variational bottleneck where the variance $\sigma(x)$ is learned directly from the input sample, optimized via the standard Evidence Lower Bound (ELBO) with a Kullback-Leibler (KL) divergence penalty. We explicitly highlight this architecture as our \textit{per-sample variance baseline}. Under extremely scarce supervision (e.g., 1\% label budgets), this framework inevitably exhibits severe likelihood blow-up; the unconstrained reconstruction objective hijacks the latent topology, forcing the network to map unstructured visual noise directly into the learned variance. This results in catastrophic posterior collapse, rendering the grounded symbolic semantics mathematically non-identifiable. 

In stark contrast, D-OCB averts this pathology through a dynamically calibrated, fixed noise floor ($\sigma_c$) alongside our Dynamic Dimension Allocation (DDA) mechanism. Instead of allowing representations to collapse, DDA intelligently redistributes latent capacity to underperforming predicates during training---specifically allocating necessary budget to resolve ambiguity in complex subspaces like ``color''---ensuring stable convergence and robust logical grounding.

\textbf{Slot VAE-NOVAE:} Finally, to ablate the impact of the generative pressure itself, we evaluate Slot VAE-NOVAE. This variant maintains the architectural topology of the Slot-VAE but completely removes the spatial broadcast decoder and the KL-divergence constraints ($\beta = 0$), collapsing the feature extraction into a purely deterministic projection.

\section{RQ1: Visual Concept Extraction and Predicate Accuracy}
\label{sec:rq1}

To evaluate the representational quality of our perception module, we measure its capability to ground continuous visual features into discrete symbolic predicates using the CLEVR and HAM10000 datasets under extreme weak supervision regimes ($1\%$ to $15\%$). Unlike traditional slot models that frequently exhibit performance collapse on complex attributes due to unconstrained reconstruction objectives hijacking the latent space topology, our Dynamic Orthogonal Concept Bottleneck (D-OCB) explicitly prevents features from dissolving into unstructured noise.

\begin{table*}[t]
\centering
\caption{Concept classification accuracy on the CLEVR dataset across varying levels of weak supervision. Results are reported as mean $\pm$ standard deviation over multiple folds.}
\label{tab:rq1_clevr_updated}
\resizebox{\textwidth}{!}{
\begin{tabular}{ll ccccc c ccccc}
\toprule
& & \multicolumn{5}{c}{\textbf{CNN Backbone}} && \multicolumn{5}{c}{\textbf{DINO Backbone}} \\
\cmidrule{3-7} \cmidrule{9-13}
\textbf{Sup.} & \textbf{Method} & Size & Shape & Color & Material & Mean && Size & Shape & Color & Material & Mean \\
\midrule
\multirow{7}{*}{\textbf{1\%}} & Vanilla+SA & 51.2 $\pm$ 0.0 & 49.3 $\pm$ 0.0 & 10.7 $\pm$ 0.0 & 62.2 $\pm$ 0.0 & 46.4 $\pm$ 0.0 && 72.3 $\pm$ 0.0 & 63.4 $\pm$ 0.0 & 15.8 $\pm$ 0.0 & 70.6 $\pm$ 0.0 & 57.3 $\pm$ 0.0 \\
 & CEM & 51.2 $\pm$ 1.3 & 33.9 $\pm$ 0.5 & 17.2 $\pm$ 6.3 & 50.9 $\pm$ 0.5 & 38.3 $\pm$ 2.1 && 73.4 $\pm$ 2.2 & 41.3 $\pm$ 3.9 & 12.3 $\pm$ 0.1 & 70.8 $\pm$ 6.6 & 49.5 $\pm$ 2.6 \\
 & ProbCBM & 48.2 $\pm$ 3.9 & 34.5 $\pm$ 2.1 & 20.9 $\pm$ 8.2 & 50.7 $\pm$ 0.7 & 38.6 $\pm$ 3.3 && 72.3 $\pm$ 4.4 & 38.4 $\pm$ 2.2 & 12.5 $\pm$ 0.3 & 67.0 $\pm$ 7.4 & 47.5 $\pm$ 3.2 \\
 & SlotFormer & 25.2 $\pm$ 3.1 & 25.0 $\pm$ 2.0 & 17.5 $\pm$ 5.1 & 50.1 $\pm$ 1.1 & 29.4 $\pm$ 1.7 && 48.0 $\pm$ 2.8 & 38.3 $\pm$ 2.8 & 17.7 $\pm$ 3.9 & 55.0 $\pm$ 3.1 & 39.8 $\pm$ 2.1 \\
 & SlotVAE-NoVAE & 52.8 $\pm$ 0.0 & 51.5 $\pm$ 0.0 & 8.9 $\pm$ 0.0 & 60.5 $\pm$ 0.0 & 48.6 $\pm$ 0.0 && 76.2 $\pm$ 0.0 & 62.6 $\pm$ 0.0 & 12.1 $\pm$ 0.0 & 75.6 $\pm$ 0.0 & 56.9 $\pm$ 0.0 \\
 & SlotVAE-KL & 62.0 $\pm$ 4.8 & 64.0 $\pm$ 6.6 & 16.9 $\pm$ 3.7 & 79.2 $\pm$ 1.4 & 55.0 $\pm$ 3.2 && 90.5 $\pm$ 6.7 & 80.7 $\pm$ 6.4 & \textbf{38.3} $\pm$ 0.2 & 93.6 $\pm$ 3.2 & 75.3 $\pm$ 3.3 \\
 & \textbf{D-OCB (Ours)} & \textbf{75.1} $\pm$ 16.1 & \textbf{67.5} $\pm$ 3.4 & \textbf{28.2} $\pm$ 5.1 & \textbf{84.5} $\pm$ 5.0 & \textbf{63.8} $\pm$ 7.1 && \textbf{94.4} $\pm$ 1.4 & \textbf{82.3} $\pm$ 8.6 & 36.7 $\pm$ 6.3 & \textbf{94.9} $\pm$ 2.3 & \textbf{77.1} $\pm$ 4.5 \\
\midrule
\multirow{5}{*}{\textbf{5\%}} & CEM & 53.5 $\pm$ 1.1 & 35.9 $\pm$ 0.6 & 24.5 $\pm$ 0.5 & 51.9 $\pm$ 1.0 & 41.5 $\pm$ 0.5 && 80.2 $\pm$ 2.9 & 48.7 $\pm$ 1.5 & 12.4 $\pm$ 0.2 & 82.7 $\pm$ 2.9 & 56.0 $\pm$ 1.0 \\
 & ProbCBM & 48.1 $\pm$ 5.2 & 34.5 $\pm$ 1.5 & 25.0 $\pm$ 0.9 & 50.9 $\pm$ 0.7 & 39.6 $\pm$ 1.6 && 74.6 $\pm$ 5.5 & 46.8 $\pm$ 4.9 & 12.6 $\pm$ 0.4 & 79.8 $\pm$ 3.9 & 53.5 $\pm$ 3.1 \\
 & SlotFormer & 26.7 $\pm$ 2.7 & 26.1 $\pm$ 3.2 & 24.8 $\pm$ 0.5 & 50.3 $\pm$ 0.8 & 32.0 $\pm$ 0.9 && 46.4 $\pm$ 3.1 & 43.9 $\pm$ 4.1 & 14.1 $\pm$ 1.8 & 63.2 $\pm$ 3.9 & 41.9 $\pm$ 2.4 \\
 & SlotVAE-KL & 85.8 $\pm$ 5.2 & 68.5 $\pm$ 4.5 & 43.0 $\pm$ 2.3 & \textbf{91.9} $\pm$ 1.6 & 72.3 $\pm$ 2.6 && 91.0 $\pm$ 7.9 & 87.3 $\pm$ 5.9 & 62.3 $\pm$ 0.1 & 96.6 $\pm$ 5.6 & 84.1 $\pm$ 4.1 \\
 & \textbf{D-OCB (Ours)} & \textbf{91.8} $\pm$ 3.3 & \textbf{81.6} $\pm$ 1.5 & \textbf{58.5} $\pm$ 2.3 & 91.5 $\pm$ 1.8 & \textbf{80.8} $\pm$ 1.0 && \textbf{97.6} $\pm$ 0.0 & \textbf{95.1} $\pm$ 1.1 & \textbf{75.4} $\pm$ 1.5 & \textbf{97.1} $\pm$ 0.1 & \textbf{91.3} $\pm$ 0.4 \\
\midrule
\multirow{7}{*}{\textbf{15\%}} & Vanilla+SA & 73.3 $\pm$ 0.0 & 66.5 $\pm$ 0.0 & 59.3 $\pm$ 0.0 & 69.7 $\pm$ 0.0 & 74.1 $\pm$ 0.0 && 77.8 $\pm$ 0.0 & 78.0 $\pm$ 0.0 & 63.9 $\pm$ 0.0 & 78.9 $\pm$ 0.0 & 70.8 $\pm$ 0.0 \\
 & CEM & 61.5 $\pm$ 2.7 & 37.0 $\pm$ 1.2 & 24.7 $\pm$ 1.1 & 53.0 $\pm$ 0.9 & 44.0 $\pm$ 1.2 && 86.2 $\pm$ 1.2 & 54.0 $\pm$ 1.5 & 12.7 $\pm$ 0.4 & 86.7 $\pm$ 0.8 & 60.0 $\pm$ 0.4 \\
 & ProbCBM & 51.4 $\pm$ 4.6 & 34.5 $\pm$ 1.1 & 25.0 $\pm$ 0.5 & 51.4 $\pm$ 0.8 & 40.6 $\pm$ 1.3 && 81.6 $\pm$ 3.7 & 48.6 $\pm$ 4.8 & 13.8 $\pm$ 1.6 & 82.7 $\pm$ 2.0 & 56.7 $\pm$ 2.5 \\
 & SlotFormer & 26.1 $\pm$ 3.8 & 25.7 $\pm$ 3.3 & 24.9 $\pm$ 0.2 & 50.0 $\pm$ 1.0 & 31.7 $\pm$ 1.2 && 47.3 $\pm$ 2.7 & 45.6 $\pm$ 4.9 & 16.7 $\pm$ 3.7 & 65.5 $\pm$ 3.6 & 43.8 $\pm$ 2.7 \\
 & SlotVAE-NoVAE & 75.5 $\pm$ 0.0 & 65.8 $\pm$ 0.0 & 61.2 $\pm$ 0.0 & 74.8 $\pm$ 0.0 & 66.5 $\pm$ 0.0 && 79.8 $\pm$ 0.0 & 78.8 $\pm$ 0.0 & 64.1 $\pm$ 0.0 & 76.2 $\pm$ 0.0 & 77.3 $\pm$ 0.0 \\
 & SlotVAE-KL & 91.6 $\pm$ 5.7 & 74.4 $\pm$ 3.7 & 52.9 $\pm$ 1.8 & 84.2 $\pm$ 2.1 & 75.8 $\pm$ 2.4 && \textbf{97.6} $\pm$ 8.2 & 92.6 $\pm$ 4.3 & 74.5 $\pm$ 0.1 & 95.9 $\pm$ 6.4 & 89.2 $\pm$ 4.0 \\
 & \textbf{D-OCB (Ours)} & \textbf{94.5} $\pm$ 1.6 & \textbf{88.6} $\pm$ 2.2 & \textbf{84.0} $\pm$ 0.7 & \textbf{93.4} $\pm$ 1.3 & \textbf{90.1} $\pm$ 0.9 && \textbf{97.6} $\pm$ 0.0 & \textbf{95.8} $\pm$ 0.3 & \textbf{84.6} $\pm$ 0.6 & \textbf{97.2} $\pm$ 0.1 & \textbf{93.8} $\pm$ 0.1 \\
\midrule
\multirow{7}{*}{\textbf{25\%}} & Vanilla+SA & 74.9 $\pm$ 0.0 & 73.3 $\pm$ 0.0 & 66.5 $\pm$ 0.0 & 70.8 $\pm$ 0.0 & 66.5 $\pm$ 0.0 && 76.8 $\pm$ 0.0 & 72.1 $\pm$ 0.0 & 66.6 $\pm$ 0.0 & 79.5 $\pm$ 0.0 & 77.6 $\pm$ 0.0 \\
 & CEM & 64.8 $\pm$ 1.8 & 38.2 $\pm$ 2.0 & 24.6 $\pm$ 1.2 & 53.3 $\pm$ 0.7 & 45.2 $\pm$ 0.8 && 88.2 $\pm$ 1.4 & 56.0 $\pm$ 0.6 & 13.5 $\pm$ 1.3 & 88.0 $\pm$ 0.7 & 61.4 $\pm$ 0.4 \\
 & ProbCBM & 52.8 $\pm$ 4.0 & 34.4 $\pm$ 1.1 & 25.0 $\pm$ 0.5 & 51.7 $\pm$ 0.6 & 41.0 $\pm$ 1.0 && 83.6 $\pm$ 3.5 & 50.1 $\pm$ 5.1 & 15.1 $\pm$ 2.2 & 84.7 $\pm$ 2.0 & 58.4 $\pm$ 2.5 \\
 & SlotFormer & 26.3 $\pm$ 3.4 & 26.0 $\pm$ 3.8 & 24.9 $\pm$ 0.3 & 49.8 $\pm$ 0.9 & 31.8 $\pm$ 1.3 && 46.6 $\pm$ 2.8 & 45.6 $\pm$ 5.1 & 17.3 $\pm$ 4.1 & 65.8 $\pm$ 3.7 & 43.8 $\pm$ 2.8 \\
 & SlotVAE-NoVAE & 79.7 $\pm$ 0.0 & 74.0 $\pm$ 0.0 & 59.5 $\pm$ 0.0 & 76.8 $\pm$ 0.0 & 74.0 $\pm$ 0.0 && 79.6 $\pm$ 0.0 & 77.4 $\pm$ 0.0 & 63.3 $\pm$ 0.0 & 75.0 $\pm$ 0.0 & 69.7 $\pm$ 0.0 \\
 & SlotVAE-KL & \textbf{98.6} $\pm$ 5.5 & \textbf{89.3} $\pm$ 3.7 & 42.0 $\pm$ 1.4 & \textbf{96.1} $\pm$ 2.3 & 75.8 $\pm$ 2.2 && 97.5 $\pm$ 8.2 & 94.4 $\pm$ 4.6 & 84.2 $\pm$ 0.1 & 96.2 $\pm$ 6.4 & 93.1 $\pm$ 4.0 \\
 & \textbf{D-OCB (Ours)} & 95.1 $\pm$ 0.9 & 89.2 $\pm$ 1.8 & \textbf{84.4} $\pm$ 0.2 & 94.4 $\pm$ 0.8 & \textbf{90.8} $\pm$ 0.5 && \textbf{97.6} $\pm$ 0.0 & \textbf{95.9} $\pm$ 0.4 & \textbf{84.7} $\pm$ 0.1 & \textbf{97.2} $\pm$ 0.1 & \textbf{93.9} $\pm$ 0.1 \\
\midrule
\multirow{7}{*}{\textbf{50\%}} & Vanilla+SA & 74.3 $\pm$ 0.0 & 66.4 $\pm$ 0.0 & 62.6 $\pm$ 0.0 & 78.0 $\pm$ 0.0 & 68.7 $\pm$ 0.0 && 80.2 $\pm$ 0.0 & 78.2 $\pm$ 0.0 & 59.8 $\pm$ 0.0 & 72.3 $\pm$ 0.0 & 69.0 $\pm$ 0.0 \\
 & CEM & 68.5 $\pm$ 1.1 & 40.5 $\pm$ 2.3 & 24.7 $\pm$ 1.1 & 54.2 $\pm$ 0.7 & 47.0 $\pm$ 0.5 && 90.5 $\pm$ 1.2 & 61.7 $\pm$ 2.5 & 17.5 $\pm$ 4.0 & 89.8 $\pm$ 0.6 & 64.9 $\pm$ 1.3 \\
 & ProbCBM & 55.1 $\pm$ 3.4 & 34.5 $\pm$ 1.3 & 25.0 $\pm$ 0.4 & 52.2 $\pm$ 0.6 & 41.7 $\pm$ 0.9 && 86.5 $\pm$ 2.5 & 53.3 $\pm$ 5.1 & 17.5 $\pm$ 3.1 & 86.4 $\pm$ 1.4 & 60.9 $\pm$ 2.2 \\
 & SlotFormer & 26.3 $\pm$ 3.4 & 26.0 $\pm$ 3.8 & 25.0 $\pm$ 0.0 & 49.8 $\pm$ 0.9 & 31.8 $\pm$ 1.3 && 46.6 $\pm$ 2.8 & 45.6 $\pm$ 5.1 & 17.3 $\pm$ 4.1 & 65.8 $\pm$ 3.7 & 43.8 $\pm$ 2.8 \\
 & SlotVAE-NoVAE & 76.2 $\pm$ 0.0 & 72.1 $\pm$ 0.0 & 63.3 $\pm$ 0.0 & 71.1 $\pm$ 0.0 & 68.1 $\pm$ 0.0 && 74.4 $\pm$ 0.0 & 71.8 $\pm$ 0.0 & 60.8 $\pm$ 0.0 & 79.9 $\pm$ 0.0 & 73.3 $\pm$ 0.0 \\
 & SlotVAE-KL & \textbf{98.5} $\pm$ 5.5 & \textbf{89.6} $\pm$ 3.7 & 69.9 $\pm$ 1.4 & \textbf{96.1} $\pm$ 2.3 & 83.5 $\pm$ 2.2 && 97.4 $\pm$ 8.2 & 94.1 $\pm$ 4.6 & \textbf{84.9} $\pm$ 0.1 & 96.1 $\pm$ 6.4 & 93.1 $\pm$ 4.0 \\
 & \textbf{D-OCB (Ours)} & 95.4 $\pm$ 1.3 & 89.1 $\pm$ 1.5 & \textbf{84.4} $\pm$ 0.3 & 94.9 $\pm$ 0.8 & \textbf{90.9} $\pm$ 0.3 && \textbf{97.6} $\pm$ 0.0 & \textbf{95.9} $\pm$ 0.4 & 84.8 $\pm$ 0.1 & \textbf{97.2} $\pm$ 0.0 & \textbf{93.9} $\pm$ 0.1 \\
\midrule
\multirow{7}{*}{\textbf{75\%}} & Vanilla+SA & 76.7 $\pm$ 0.0 & 66.1 $\pm$ 0.0 & 68.6 $\pm$ 0.0 & 71.7 $\pm$ 0.0 & 68.8 $\pm$ 0.0 && 80.2 $\pm$ 0.0 & 75.5 $\pm$ 0.0 & 60.8 $\pm$ 0.0 & 72.3 $\pm$ 0.0 & 69.8 $\pm$ 0.0 \\
 & CEM & 70.8 $\pm$ 1.2 & 43.1 $\pm$ 2.6 & 24.8 $\pm$ 1.0 & 54.9 $\pm$ 0.6 & 48.4 $\pm$ 0.4 && 91.3 $\pm$ 0.9 & 66.1 $\pm$ 1.9 & 23.1 $\pm$ 4.0 & 90.6 $\pm$ 0.5 & 67.8 $\pm$ 1.0 \\
 & ProbCBM & 56.5 $\pm$ 2.8 & 34.7 $\pm$ 1.2 & 25.0 $\pm$ 0.3 & 52.6 $\pm$ 0.8 & 42.2 $\pm$ 0.7 && 88.1 $\pm$ 2.1 & 55.0 $\pm$ 5.1 & 19.4 $\pm$ 3.6 & 87.6 $\pm$ 1.5 & 62.5 $\pm$ 2.1 \\
 & SlotFormer & 26.3 $\pm$ 3.4 & 26.0 $\pm$ 3.8 & 25.0 $\pm$ 0.0 & 49.8 $\pm$ 0.9 & 31.8 $\pm$ 1.3 && 46.6 $\pm$ 2.8 & 45.6 $\pm$ 5.1 & 17.3 $\pm$ 4.1 & 65.8 $\pm$ 3.7 & 43.8 $\pm$ 2.8 \\
 & SlotVAE-NoVAE & 74.5 $\pm$ 0.0 & 73.4 $\pm$ 0.0 & 68.3 $\pm$ 0.0 & 71.7 $\pm$ 0.0 & 70.4 $\pm$ 0.0 && 74.2 $\pm$ 0.0 & 78.0 $\pm$ 0.0 & 59.8 $\pm$ 0.0 & 78.8 $\pm$ 0.0 & 73.3 $\pm$ 0.0 \\
 & SlotVAE-KL & \textbf{98.7} $\pm$ 5.5 & \textbf{89.5} $\pm$ 3.7 & 82.8 $\pm$ 1.4 & \textbf{95.1} $\pm$ 2.3 & 89.0 $\pm$ 2.2 && \textbf{98.2} $\pm$ 8.2 & 94.9 $\pm$ 4.6 & 87.4 $\pm$ 0.1 & 96.9 $\pm$ 6.4 & 94.3 $\pm$ 4.0 \\
 & \textbf{D-OCB (Ours)} & 95.4 $\pm$ 1.3 & 89.1 $\pm$ 1.5 & \textbf{84.4} $\pm$ 0.3 & 95.0 $\pm$ 0.9 & \textbf{91.0} $\pm$ 0.3 && 97.6 $\pm$ 0.0 & \textbf{95.9} $\pm$ 0.4 & \textbf{94.8} $\pm$ 0.1 & \textbf{97.2} $\pm$ 0.0 & \textbf{95.5} $\pm$ 0.1 \\
\bottomrule
\end{tabular}
}
\end{table*}

\begin{table}[t]
\centering
\small
\setlength{\tabcolsep}{5pt}
\caption{Concept classification accuracy of \textbf{D-OCB} on the HAM10000 dataset across varying levels of weak supervision. Results are reported as mean $\pm$ standard deviation.}
\label{tab:rq1_ham_concepts}
\begin{tabular}{@{}ll ccccc@{}}
\toprule
\textbf{Backbone} & \textbf{Sup. (\%)} & \textbf{Dx} & \textbf{Localization} & \textbf{Sex} & \textbf{Age} & \textbf{Mean} \\
\midrule
\multirow{6}{*}{\textbf{CNN}} 
 & 1 & $55.3 \pm 0.0$ & $84.2 \pm 0.0$ & $22.1 \pm 0.0$ & $50.0 \pm 0.0$ & $52.9 \pm 0.0$ \\
 & 5 & $55.0 \pm 0.0$ & $84.4 \pm 0.0$ & $60.6 \pm 0.0$ & $61.7 \pm 0.0$ & $65.4 \pm 0.0$ \\
 & 15 & $59.3 \pm 0.0$ & $86.0 \pm 0.0$ & $64.3 \pm 0.0$ & $64.6 \pm 0.0$ & $68.6 \pm 0.0$ \\
 & 25 & $60.0 \pm 1.9$ & $85.8 \pm 0.6$ & $63.2 \pm 2.4$ & $65.5 \pm 1.5$ & $68.6 \pm 1.3$ \\
 & 50 & $63.0 \pm 1.1$ & $87.5 \pm 0.8$ & $69.2 \pm 1.6$ & $67.5 \pm 1.2$ & $71.8 \pm 1.1$ \\
 & 75 & $65.3 \pm 1.7$ & $89.1 \pm 0.7$ & $72.6 \pm 2.5$ & $68.7 \pm 2.5$ & $73.9 \pm 1.8$ \\
\midrule
\multirow{6}{*}{\textbf{DINO}} 
 & 1 & $61.8 \pm 0.0$ & $82.5 \pm 0.0$ & $67.4 \pm 0.0$ & $60.2 \pm 0.0$ & $68.0 \pm 0.0$ \\
 & 5 & $64.4 \pm 0.0$ & $78.5 \pm 0.0$ & $69.4 \pm 0.0$ & $69.1 \pm 0.0$ & $70.4 \pm 0.0$ \\
 & 15 & $64.2 \pm 0.0$ & $85.5 \pm 0.0$ & $72.8 \pm 0.0$ & $70.9 \pm 0.0$ & $73.4 \pm 0.0$ \\
 & 25 & $67.2 \pm 1.5$ & $86.7 \pm 0.1$ & $73.1 \pm 1.3$ & $72.0 \pm 1.2$ & $74.8 \pm 0.1$ \\
 & 50 & $69.7 \pm 1.2$ & $89.3 \pm 1.9$ & $79.5 \pm 2.6$ & $75.5 \pm 1.3$ & $78.5 \pm 1.7$ \\
 & 75 & $72.1 \pm 2.3$ & $90.2 \pm 0.3$ & $81.2 \pm 4.1$ & $77.3 \pm 1.3$ & $80.2 \pm 2.0$ \\
\bottomrule
\end{tabular}
\end{table}
 By discarding the per-sample learned variance and enforcing strict semantic boundaries through a cross-correlation orthogonal constraint and dynamic dimensionality allocation, D-OCB maintains robust cluster separability even when concept annotations are severely restricted. As shown in Table~\ref{tab:rq1_ham_concepts} and Table \ref{tab:rq1_clevr_updated}, D-OCB significantly outperforms baselines by dynamically assigning dimensions to concepts lacking accuracy, achieving highly accurate and well-grounded visual predicates (e.g., maintaining a $73.4\%$ mean accuracy on HAM10000 with a DINO backbone at just $15\%$ supervision) that preserve stable concept alignment without the need for manual hyperparameter tuning.

\section{RQ2: Framework-Agnostic Logical Reasoning Evaluation}
\label{sec:rq2}

\begin{table}[tb]
\centering
\tiny
\caption{Downstream evaluation rules for the CLEVR and HAM10000 datasets.}
\label{tab:rules}
\begin{tabular}{@{}c@{\,\,}l@{\,\,} l@{\,\,} l@{}}
\toprule
& \textbf{ID} & \textbf{Rule Type} & \textbf{Description} \\ 
\midrule
\multirow{5}{*}{\rotatebox{90}{\textit{CLEVR}}}&
\textbf{R1} & 4-way Conjunction  & The scene contains an object that is simultaneously large, red, metal, and a cube. \\&
\textbf{R2} & 3-way Conjunction  & The scene contains a small rubber sphere. \\&
\textbf{R3} & Spatial Relation   & A blue cylinder is spatially to the left of a red object (evaluated via 3D x-coordinates). \\&
\textbf{R4} & Cardinality        & The scene contains $\ge 2$ metal objects; requires the model to count distinct slots satisfying this constraint. \\&
\textbf{R5} & Proximity Relation & A large green object is within a continuous Euclidean distance threshold ($\le 2.5$) of a yellow object. \\
\midrule
\multirow{3}{*}{\rotatebox{90}{\textit{HAM}}}&
\textbf{R1} & Simple         & The lesion is malignant. \\&
\textbf{R2} & Disjunctive    & The lesion is melanoma (mel) or basal cell carcinoma (bcc). \\&
\textbf{R3} & Conjunctive    & The lesion is melanoma located on the back (a diagnosis and location conjunction). \\
\bottomrule
\end{tabular}
\end{table}

Having verified predicate grounding accuracy, we investigate whether these extracted visual atoms are sufficiently stable to drive downstream symbolic reasoning engines across complex compositional rules on CLEVR (evaluating conjunctions, spatial relations, and cardinality constraints up to R5) and HAM10000 (evaluating simple, disjunctive, and conjunctive diagnostic rules). By strictly isolating the perception stage, we export the frozen valuation tensors generated by D-OCB under extremely low supervision to independent offline logic frameworks, including Differentiable Inductive Logic Programming (DILP) \cite{DBLP:journals/corr/abs-2103-01719}, Decision Trees, Bayesian Networks, and the Neural-Symbolic Concept Learner (NS-CL). To rigorously test this, we export the frozen valuation tensors generated by D-OCB to evaluate specific compositional and diagnostic rules, as detailed in \autoref{tab:rules}.

The empirical results  in Table \ref{tab:merged_ham_reasoning} and Table \ref{tab:rq2_detailed_reasoning_compact} substantiate the computational advantages of separating visual parsing from programmatic reasoning, as end-to-end architectures suffer heavily from Posterior collapse  where unconstrained error signals encourage the neural backbone to cross-contaminate concept boundaries. Crucially, by structurally blocking this  bleed, the softly-valued probabilistic margins preserved within D-OCB's causal subspaces allow downstream statistical reasoners to establish optimal decision boundaries, consistently matching and occasionally surpassing the performance of static, binary ground-truth Oracle annotations.

\begin{figure}
    \centering
    \includegraphics[width=0.5\linewidth]{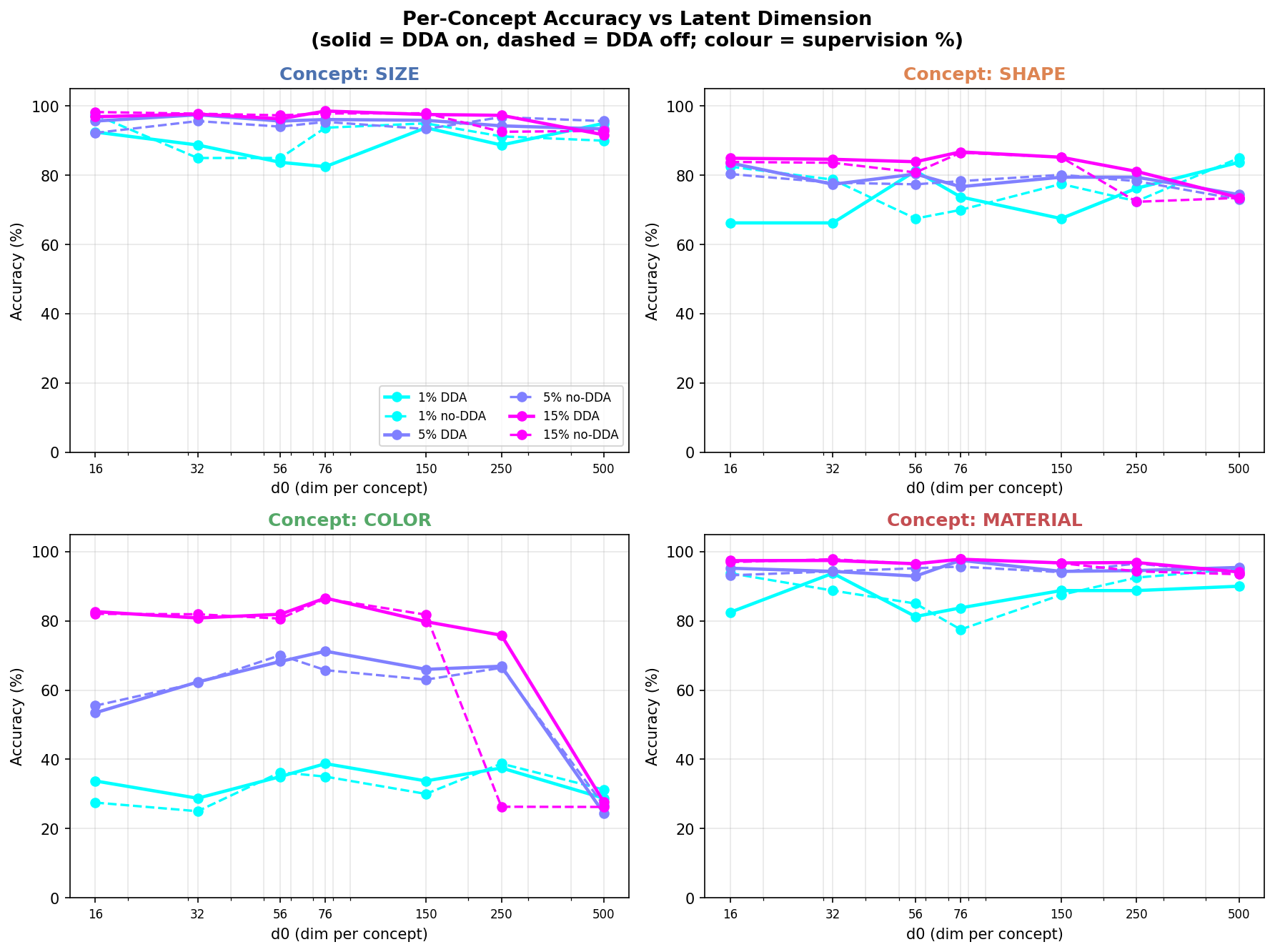}
    \caption{Clevr Hans 7 per concept accuracy}
    \label{fig:DDA HANS7}
\end{figure}

\begin{table*}[t]
\centering
\footnotesize
\setlength{\tabcolsep}{3pt}
\renewcommand{\arraystretch}{0.9}
\caption{Detailed downstream reasoning accuracy (\%) across five compositional rules on the CLEVR dataset. The data is transposed to consolidate all supervisions into columns, drastically reducing the vertical footprint. Standard deviations are strictly $0.0$ across tested seeds and are thus omitted.}
\label{tab:rq2_detailed_reasoning_compact}
\resizebox{\textwidth}{!}{
\begin{tabular}{ll *{20}{c}}
\toprule
\multirow{2}{*}{\textbf{Rule (Backbone)}} & \multirow{2}{*}{\textbf{Perception Model}} & \multicolumn{5}{c}{\textbf{Decision Tree (DT)}} & \multicolumn{5}{c}{\textbf{Bayes Net (BN)}} & \multicolumn{5}{c}{\textbf{NS-CL}} & \multicolumn{5}{c}{\textbf{DILP}} \\
\cmidrule(lr){3-7} \cmidrule(lr){8-12} \cmidrule(lr){13-17} \cmidrule(lr){18-22}
& & 1\% & 15\% & 25\% & 50\% & 75\% & 1\% & 15\% & 25\% & 50\% & 75\% & 1\% & 15\% & 25\% & 50\% & 75\% & 1\% & 15\% & 25\% & 50\% & 75\% \\
\midrule
\multirow{4}{*}{\textbf{R1 (CNN)}}
& \textbf{D-OCB (Ours)} & \textbf{69.6} & \textbf{58.8} & 65.9 & 63.7 & 64.5 & \textbf{94.2} & \textbf{57.5} & 58.4 & 57.5 & 58.5 & \textbf{57.6} & \textbf{55.7} & 56.6 & 55.7 & 56.1 & 51.8 & \textbf{99.6} & \textbf{99.7} & \textbf{99.8} & \textbf{99.7} \\
& SlotVAE-KL & 44.0 & 9.6 & \textbf{68.7} & 61.3 & 58.1 & 33.0 & 9.2 & \textbf{91.6} & \textbf{94.0} & \textbf{93.1} & 5.8 & 5.8 & 28.0 & \textbf{75.3} & \textbf{76.8} & 19.1 & 94.2 & 74.3 & 88.1 & 89.0 \\
& SlotVAE-NoVAE & 46.5 & 33.1 & 38.1 & \textbf{67.9} & \textbf{65.6} & 22.0 & 21.7 & 22.8 & 25.0 & 24.5 & 53.3 & 48.2 & \textbf{60.6} & 63.3 & 62.5 & \textbf{74.1} & 79.6 & 91.8 & 98.8 & 98.8 \\
& Vanilla+SA & 30.1 & 40.0 & 49.0 & 44.8 & 44.9 & 17.0 & 22.1 & 23.5 & 23.3 & 24.2 & 52.3 & 52.4 & 54.6 & 54.7 & 67.3 & 51.6 & 77.0 & 88.7 & 84.3 & 94.3 \\
\midrule
\multirow{4}{*}{\textbf{R1 (DINO)}}
& \textbf{D-OCB (Ours)} & 60.7 & 59.8 & \textbf{65.9} & 69.8 & 57.1 & 22.7 & 56.6 & 57.1 & 56.9 & 56.7 & 57.6 & 55.7 & 55.3 & 55.3 & 55.7 & 98.7 & \textbf{99.7} & \textbf{99.6} & \textbf{99.7} & \textbf{99.7} \\
& SlotVAE-KL & 59.2 & \textbf{71.1} & 61.1 & \textbf{70.1} & \textbf{67.4} & \textbf{93.7} & \textbf{58.2} & \textbf{94.1} & \textbf{94.0} & \textbf{94.2} & \textbf{71.9} & 56.6 & \textbf{71.3} & \textbf{70.9} & \textbf{69.9} & 87.2 & 99.6 & 88.6 & 88.4 & 88.6 \\
& SlotVAE-NoVAE & 56.4 & 57.4 & 58.9 & 57.1 & 57.2 & 57.5 & 56.4 & 57.6 & 56.6 & 57.3 & 57.7 & \textbf{56.7} & 57.3 & 55.2 & 55.7 & \textbf{98.8} & 99.2 & 99.5 & 99.6 & 99.6 \\
& Vanilla+SA & \textbf{64.7} & 58.8 & 60.2 & 62.3 & 56.5 & 25.3 & 56.5 & 57.6 & 55.7 & 56.8 & 53.5 & 56.4 & 57.7 & 55.6 & 55.9 & 96.2 & 99.4 & 99.3 & 99.1 & 99.1 \\
\midrule
\multirow{4}{*}{\textbf{R2 (CNN)}}
& \textbf{D-OCB (Ours)} & 60.0 & 59.8 & 60.9 & \textbf{61.5} & 59.3 & \textbf{56.0} & \textbf{56.9} & 57.2 & \textbf{57.3} & 57.4 & 58.2 & \textbf{61.1} & 62.1 & \textbf{61.9} & 61.9 & 46.1 & \textbf{95.4} & \textbf{95.4} & \textbf{96.3} & 95.0 \\
& SlotVAE-KL & \textbf{64.4} & 46.3 & \textbf{63.7} & 60.7 & \textbf{68.8} & 54.1 & 42.9 & 56.5 & 55.6 & \textbf{59.9} & 49.0 & 42.9 & 61.9 & 61.6 & \textbf{64.2} & 51.1 & 42.9 & 57.6 & 57.9 & 60.8 \\
& SlotVAE-NoVAE & 59.0 & 58.1 & 59.4 & 58.9 & 61.0 & 51.2 & 56.6 & \textbf{57.8} & 57.2 & 57.6 & \textbf{61.0} & 56.7 & 58.8 & 61.6 & 62.6 & 61.3 & 77.9 & 87.0 & 93.4 & 95.4 \\
& Vanilla+SA & 58.2 & \textbf{60.6} & 59.2 & 61.3 & 61.3 & 52.4 & 56.5 & 56.7 & 57.2 & 57.6 & \textbf{61.0} & 60.5 & \textbf{62.5} & 58.8 & 59.1 & \textbf{66.5} & 78.9 & 83.6 & 90.6 & \textbf{95.9} \\
\midrule
\multirow{4}{*}{\textbf{R2 (DINO)}}
& \textbf{D-OCB (Ours)} & 60.0 & 61.9 & 58.7 & 63.3 & 62.9 & 57.3 & 57.4 & 57.3 & 57.0 & 57.4 & 61.8 & 62.1 & 62.2 & 61.7 & 61.8 & 91.3 & 95.0 & \textbf{96.5} & \textbf{96.3} & \textbf{97.3} \\
& SlotVAE-KL & \textbf{67.0} & \textbf{64.1} & \textbf{69.5} & \textbf{68.8} & \textbf{68.9} & \textbf{68.1} & \textbf{57.8} & \textbf{69.4} & \textbf{70.1} & \textbf{67.0} & \textbf{65.9} & \textbf{62.3} & \textbf{69.7} & \textbf{69.0} & \textbf{67.6} & 59.0 & 96.2 & 69.5 & 70.7 & 61.6 \\
& SlotVAE-NoVAE & 60.5 & 59.7 & 58.8 & 57.3 & 58.5 & 57.8 & 57.6 & 57.7 & 57.5 & 57.3 & 62.1 & 61.6 & 62.3 & 62.0 & 61.9 & \textbf{94.8} & \textbf{96.6} & 95.5 & 96.5 & 97.1 \\
& Vanilla+SA & 61.1 & 62.6 & 59.5 & 61.2 & 58.9 & 57.5 & 57.5 & 57.8 & 57.5 & 57.5 & 63.5 & 62.2 & 62.0 & 62.1 & 61.6 & 91.5 & 95.4 & 93.6 & 96.0 & 95.5 \\
\midrule
\multirow{4}{*}{\textbf{R3 (CNN)}}
& \textbf{D-OCB (Ours)} & \textbf{50.9} & \textbf{77.6} & \textbf{76.3} & \textbf{76.6} & 76.5 & 91.8 & 78.6 & 79.6 & 78.9 & 78.6 & 54.0 & \textbf{76.9} & \textbf{77.1} & \textbf{77.2} & 77.0 & \textbf{73.9} & \textbf{94.9} & \textbf{94.8} & \textbf{95.3} & \textbf{95.1} \\
& SlotVAE-KL & 43.9 & 14.2 & 52.0 & 66.4 & \textbf{76.9} & 46.2 & 14.2 & 88.8 & 91.4 & 91.8 & \textbf{84.2} & 11.4 & 42.7 & 72.1 & 72.4 & 34.1 & 8.0 & 51.9 & 81.2 & 82.7 \\
& SlotVAE-NoVAE & 46.8 & 43.5 & 59.9 & 66.6 & 72.2 & \textbf{92.0} & \textbf{92.0} & 91.0 & 81.8 & 82.0 & 53.6 & 48.3 & 62.5 & 75.4 & \textbf{77.6} & 57.0 & 53.2 & 84.5 & 93.1 & 93.8 \\
& Vanilla+SA & 43.8 & 45.3 & 42.0 & 42.8 & 46.0 & \textbf{92.0} & \textbf{92.0} & \textbf{92.0} & \textbf{92.0} & \textbf{83.6} & 56.1 & 54.4 & 57.7 & 54.8 & 59.0 & 49.7 & 59.2 & 76.7 & 68.9 & 84.9 \\
\midrule
\multirow{4}{*}{\textbf{R3 (DINO)}}
& \textbf{D-OCB (Ours)} & 60.5 & 73.2 & 74.3 & 75.3 & 73.6 & 72.8 & 76.4 & 76.1 & 76.8 & 76.8 & 68.6 & 74.8 & 74.2 & 75.2 & 75.4 & 89.2 & \textbf{94.0} & 94.3 & 94.8 & 94.7 \\
& SlotVAE-KL & 62.0 & 72.2 & 60.0 & 65.1 & 65.1 & \textbf{92.0} & 78.7 & \textbf{92.0} & \textbf{92.0} & \textbf{92.0} & 53.9 & \textbf{76.5} & 65.0 & 62.3 & 63.6 & 52.2 & 92.9 & 75.9 & 74.6 & 75.9 \\
& SlotVAE-NoVAE & \textbf{67.7} & \textbf{75.4} & \textbf{76.2} & \textbf{76.8} & \textbf{74.4} & 78.6 & \textbf{78.9} & 79.2 & 78.3 & 78.4 & \textbf{74.6} & 76.3 & \textbf{77.2} & \textbf{77.0} & \textbf{76.3} & \textbf{93.2} & 93.9 & \textbf{94.8} & \textbf{95.5} & \textbf{95.4} \\
& Vanilla+SA & 55.4 & 70.1 & 71.0 & 67.5 & 69.1 & 78.3 & 77.2 & 80.4 & 75.8 & 78.5 & 59.7 & 74.1 & 74.8 & 73.8 & 74.3 & 80.1 & 93.9 & 92.9 & 93.3 & 92.7 \\
\midrule
\multirow{4}{*}{\textbf{R4 (CNN)}}
& \textbf{D-OCB (Ours)} & \textbf{79.9} & 81.1 & \textbf{81.6} & \textbf{80.1} & 74.0 & \textbf{88.9} & 87.8 & 87.9 & 87.8 & 87.8 & 84.9 & 85.0 & 85.0 & 85.0 & \textbf{85.1} & 89.3 & \textbf{99.6} & \textbf{99.6} & \textbf{99.5} & \textbf{99.5} \\
& SlotVAE-KL & 77.4 & \textbf{84.0} & 78.5 & 76.9 & \textbf{80.1} & 85.2 & 84.9 & 85.3 & 87.5 & \textbf{88.6} & 84.9 & 84.9 & 84.9 & 84.9 & 84.9 & 84.9 & 84.9 & 89.1 & 88.2 & 89.0 \\
& SlotVAE-NoVAE & 72.2 & 69.2 & 77.3 & 78.9 & 75.2 & 87.2 & \textbf{88.5} & 88.1 & 88.0 & 88.1 & \textbf{85.0} & \textbf{85.4} & 85.1 & 85.0 & 85.0 & \textbf{94.5} & 97.2 & 98.8 & 99.4 & 99.2 \\
& Vanilla+SA & 76.0 & 73.7 & 76.3 & 76.4 & 73.2 & 86.9 & 88.2 & \textbf{88.5} & \textbf{88.2} & 87.9 & 84.9 & 85.3 & \textbf{85.3} & \textbf{85.2} & 85.0 & 93.4 & 97.5 & 98.0 & 98.8 & 99.4 \\
\midrule
\multirow{4}{*}{\textbf{R4 (DINO)}}
& \textbf{D-OCB (Ours)} & \textbf{80.7} & 71.8 & 74.9 & \textbf{85.0} & 79.3 & 87.6 & \textbf{88.0} & 87.7 & 87.8 & 87.7 & 84.9 & \textbf{85.0} & \textbf{85.0} & \textbf{85.0} & \textbf{85.0} & 97.0 & \textbf{99.5} & \textbf{99.9} & \textbf{99.7} & 99.6 \\
& SlotVAE-KL & 79.9 & \textbf{81.8} & \textbf{81.7} & 82.5 & \textbf{83.2} & 84.9 & 87.9 & \textbf{88.7} & \textbf{90.0} & \textbf{89.9} & 84.9 & \textbf{85.0} & 84.9 & 84.9 & 84.9 & 87.0 & 99.0 & 88.8 & 90.1 & 90.4 \\
& SlotVAE-NoVAE & 72.0 & 72.5 & 75.9 & 67.1 & 68.3 & \textbf{88.3} & 87.8 & 87.8 & 87.9 & 87.7 & \textbf{85.0} & \textbf{85.0} & \textbf{85.0} & \textbf{85.0} & \textbf{85.0} & \textbf{98.5} & 99.3 & 99.7 & \textbf{99.7} & \textbf{99.7} \\
& Vanilla+SA & 74.0 & 76.9 & 78.1 & 76.6 & 70.2 & 87.8 & \textbf{88.0} & 88.0 & 88.1 & 87.9 & 84.9 & \textbf{85.0} & \textbf{85.0} & \textbf{85.0} & \textbf{85.0} & 97.6 & 99.4 & 99.2 & 99.2 & 99.2 \\
\midrule
\multirow{4}{*}{\textbf{R5 (CNN)}}
& \textbf{D-OCB (Ours)} & \textbf{54.9} & \textbf{76.8} & \textbf{75.3} & \textbf{77.6} & \textbf{74.7} & \textbf{94.3} & \textbf{77.0} & \textbf{77.6} & \textbf{77.3} & 76.9 & \textbf{56.3} & \textbf{76.5} & \textbf{76.9} & \textbf{76.6} & \textbf{76.1} & \textbf{74.2} & \textbf{95.5} & \textbf{96.1} & \textbf{96.2} & \textbf{96.2} \\
& SlotVAE-KL & 51.8 & 12.2 & 69.9 & 73.4 & 73.9 & 46.2 & 12.2 & \textbf{94.3} & \textbf{94.1} & \textbf{94.1} & 5.7 & 12.1 & 33.1 & 76.3 & 72.8 & 36.0 & 5.7 & 74.5 & 88.8 & 89.1 \\
& SlotVAE-NoVAE & 53.3 & 37.6 & 64.1 & 68.3 & 72.8 & \textbf{94.3} & 9.0 & 90.5 & 49.2 & 55.2 & 54.2 & 45.1 & 66.2 & 75.9 & 76.9 & 67.7 & 65.3 & 82.9 & 93.4 & 92.7 \\
& Vanilla+SA & 48.6 & 39.3 & 44.2 & 52.2 & 69.3 & 8.7 & 9.0 & 8.8 & 8.5 & 8.5 & 56.6 & 49.1 & 59.0 & 60.0 & 63.6 & 60.0 & 67.7 & 79.1 & 83.5 & 86.0 \\
\midrule
\multirow{4}{*}{\textbf{R5 (DINO)}}
& \textbf{D-OCB (Ours)} & \textbf{77.4} & 74.6 & \textbf{77.5} & \textbf{76.6} & \textbf{76.5} & 37.6 & 77.1 & 76.6 & 76.5 & 76.6 & \textbf{77.2} & 76.6 & 76.2 & 75.9 & 76.1 & 93.5 & \textbf{96.4} & \textbf{96.4} & 96.3 & \textbf{96.5} \\
& SlotVAE-KL & 59.0 & 74.9 & 65.2 & 63.2 & 68.1 & \textbf{94.3} & \textbf{78.5} & \textbf{94.3} & \textbf{94.3} & \textbf{94.3} & 47.3 & \textbf{78.4} & 66.2 & 65.5 & 67.1 & 73.2 & 89.0 & 86.4 & 86.2 & 88.5 \\
& SlotVAE-NoVAE & 67.6 & 75.1 & 76.2 & 73.5 & 74.4 & 64.5 & 76.7 & 77.5 & 76.8 & 76.6 & 77.4 & 77.4 & \textbf{77.2} & 75.8 & 76.1 & \textbf{94.2} & 95.2 & 95.9 & \textbf{96.7} & 96.2 \\
& Vanilla+SA & 55.6 & \textbf{75.4} & 75.8 & 74.8 & 76.6 & \textbf{94.3} & 77.4 & 76.1 & 76.4 & 77.0 & 64.8 & 77.4 & 77.0 & \textbf{76.2} & \textbf{77.1} & 87.6 & 95.7 & 95.9 & 95.1 & 95.9 \\
\bottomrule
\end{tabular}
}
\end{table*}

\begin{table*}[t]
\centering
\caption{Quantitative reasoning comparison of \textbf{D-OCB} against Slot-VAE and Vanilla CNN/DINO baselines on the HAM10000 dataset (RQ2). Results are reported in percentage (\%) for three clinical rules (R1, R2, R3) across varying backbones, supervision levels, and reasoning frameworks. Standard deviations $< 0.001$ are denoted as $0.0$. Baseline F1 scores have been scaled to percentages for direct comparability. Missing baseline evaluations are denoted by `-'.}
\label{tab:merged_ham_reasoning}
\resizebox{\textwidth}{!}{%
\renewcommand{\arraystretch}{1.1}
\setlength{\tabcolsep}{4pt}
\begin{tabular}{@{}lll ccc ccc ccc ccc@{}}
\toprule
\multirow{2}{*}{\textbf{BB}} & \multirow{2}{*}{\textbf{Sup\%}} & \multirow{2}{*}{\textbf{Model}} & \multicolumn{3}{c}{\textbf{DILP (\%)}} & \multicolumn{3}{c}{\textbf{Decision Tree (\%)}} & \multicolumn{3}{c}{\textbf{Bayes Net (\%)}} & \multicolumn{3}{c}{\textbf{NS-CL (\%)}} \\
\cmidrule(lr){4-6} \cmidrule(lr){7-9} \cmidrule(lr){10-12} \cmidrule(lr){13-15}
& & & \textbf{R1} & \textbf{R2} & \textbf{R3} & \textbf{R1} & \textbf{R2} & \textbf{R3} & \textbf{R1} & \textbf{R2} & \textbf{R3} & \textbf{R1} & \textbf{R2} & \textbf{R3} \\
\midrule
\multirow{18}{*}{CNN} 
& \multirow{3}{*}{1} 
  & Vanilla & - & - & - & $12.7 \pm 2.0$ & $0.0 \pm 0.0$ & $0.0 \pm 0.0$ & $35.9 \pm 3.0$ & $21.3 \pm 2.0$ & $0.0 \pm 0.0$ & $21.6 \pm 0.0$ & $16.4 \pm 0.0$ & $0.0 \pm 0.0$ \\
& & S-VAE   & - & - & - & $0.0 \pm 0.0$ & $0.0 \pm 0.0$ & $0.0 \pm 0.0$ & $15.6 \pm 1.0$ & $18.6 \pm 1.0$ & $2.6 \pm 1.0$ & $0.0 \pm 0.0$ & $0.0 \pm 0.0$ & $0.0 \pm 0.0$ \\
& & \textbf{D-OCB} & $\textbf{31.8} \pm 0.0$ & $\textbf{34.8} \pm 0.0$ & $\textbf{86.6} \pm 0.0$ & $\textbf{59.7} \pm 0.0$ & $\textbf{69.2} \pm 0.0$ & $\textbf{72.6} \pm 0.0$ & $\textbf{76.6} \pm 0.0$ & $\textbf{82.6} \pm 0.0$ & $\textbf{92.5} \pm 0.0$ & $\textbf{52.7} \pm 0.0$ & $\textbf{60.2} \pm 0.0$ & $\textbf{92.5} \pm 0.0$ \\
\cmidrule(l){2-15}

& \multirow{3}{*}{5} 
  & Vanilla & - & - & - & - & - & - & - & - & - & - & - & - \\
& & S-VAE   & - & - & - & - & - & - & - & - & - & - & - & - \\
& & \textbf{D-OCB} & $\textbf{72.1} \pm 0.0$ & $\textbf{72.6} \pm 0.0$ & $\textbf{88.6} \pm 0.0$ & $\textbf{69.2} \pm 0.0$ & $\textbf{69.7} \pm 0.0$ & $\textbf{70.1} \pm 0.0$ & $\textbf{83.1} \pm 0.0$ & $\textbf{75.6} \pm 0.0$ & $\textbf{94.5} \pm 0.0$ & $\textbf{72.1} \pm 0.0$ & $\textbf{69.2} \pm 0.0$ & $\textbf{85.6} \pm 0.0$ \\
\cmidrule(l){2-15}

& \multirow{3}{*}{15} 
  & Vanilla & - & - & - & $13.9 \pm 2.0$ & $5.9 \pm 1.0$ & $0.0 \pm 0.0$ & $37.8 \pm 1.0$ & $24.4 \pm 1.0$ & $0.0 \pm 0.0$ & $31.8 \pm 0.0$ & $6.5 \pm 0.0$ & $2.8 \pm 0.0$ \\
& & S-VAE   & - & - & - & $27.2 \pm 3.0$ & $15.4 \pm 3.0$ & $0.4 \pm 1.0$ & $30.0 \pm 1.0$ & $18.5 \pm 1.0$ & $0.0 \pm 0.0$ & $43.3 \pm 0.0$ & $37.9 \pm 0.0$ & $8.8 \pm 0.0$ \\
& & \textbf{D-OCB} & $\textbf{69.7} \pm 0.0$ & $\textbf{72.6} \pm 0.0$ & $\textbf{80.1} \pm 0.0$ & $\textbf{63.2} \pm 0.0$ & $\textbf{58.2} \pm 0.0$ & $\textbf{72.1} \pm 0.0$ & $\textbf{84.6} \pm 0.0$ & $\textbf{82.6} \pm 0.0$ & $\textbf{95.0} \pm 0.0$ & $\textbf{69.2} \pm 0.0$ & $\textbf{64.2} \pm 0.0$ & $\textbf{84.1} \pm 0.0$ \\
\cmidrule(l){2-15}

& \multirow{3}{*}{25} 
  & Vanilla & - & - & - & $5.4 \pm 2.0$ & $1.2 \pm 1.0$ & $0.0 \pm 0.0$ & $11.0 \pm 1.0$ & $0.0 \pm 0.0$ & $0.0 \pm 0.0$ & $33.1 \pm 0.0$ & $19.7 \pm 0.0$ & $7.0 \pm 0.0$ \\
& & S-VAE   & - & - & - & $40.2 \pm 3.0$ & $21.5 \pm 4.0$ & $0.0 \pm 0.0$ & $36.4 \pm 3.0$ & $23.4 \pm 2.0$ & $0.0 \pm 0.0$ & $48.6 \pm 0.0$ & $42.7 \pm 0.0$ & $25.7 \pm 0.0$ \\
& & \textbf{D-OCB} & $\textbf{76.0} \pm 2.4$ & $\textbf{74.0} \pm 2.2$ & $\textbf{87.6} \pm 6.8$ & $\textbf{71.1} \pm 5.2$ & $\textbf{63.5} \pm 2.4$ & $\textbf{80.3} \pm 2.3$ & $\textbf{82.4} \pm 1.3$ & $\textbf{82.8} \pm 2.6$ & $\textbf{94.5} \pm 1.1$ & $\textbf{69.2} \pm 2.5$ & $\textbf{63.3} \pm 2.7$ & $\textbf{87.7} \pm 3.0$ \\
\cmidrule(l){2-15}

& \multirow{3}{*}{50} 
  & Vanilla & - & - & - & $11.8 \pm 2.0$ & $7.4 \pm 3.0$ & $0.0 \pm 0.0$ & $27.3 \pm 2.0$ & $25.5 \pm 2.0$ & $0.0 \pm 0.0$ & $31.1 \pm 0.0$ & $22.7 \pm 0.0$ & $6.8 \pm 0.0$ \\
& & S-VAE   & - & - & - & $40.2 \pm 2.0$ & $18.2 \pm 1.0$ & $0.0 \pm 0.0$ & $49.4 \pm 1.0$ & $36.8 \pm 1.0$ & $0.0 \pm 0.0$ & $50.9 \pm 0.0$ & $43.5 \pm 0.0$ & $18.9 \pm 0.0$ \\
& & \textbf{D-OCB} & $\textbf{83.7} \pm 3.8$ & $\textbf{81.6} \pm 0.8$ & $\textbf{91.2} \pm 1.0$ & $\textbf{73.3} \pm 4.9$ & $\textbf{74.0} \pm 3.3$ & $\textbf{82.3} \pm 4.0$ & $\textbf{85.6} \pm 1.5$ & $\textbf{83.4} \pm 2.4$ & $\textbf{92.9} \pm 2.8$ & $\textbf{74.8} \pm 3.0$ & $\textbf{68.5} \pm 2.1$ & $\textbf{88.7} \pm 1.2$ \\
\cmidrule(l){2-15}

& \multirow{3}{*}{75} 
  & Vanilla & - & - & - & $4.2 \pm 2.0$ & $1.2 \pm 1.0$ & $0.0 \pm 0.0$ & $19.5 \pm 2.0$ & $14.4 \pm 2.0$ & $0.0 \pm 0.0$ & $32.0 \pm 0.0$ & $25.2 \pm 0.0$ & $7.9 \pm 0.0$ \\
& & S-VAE   & - & - & - & $40.2 \pm 2.0$ & $18.2 \pm 1.0$ & $0.0 \pm 0.0$ & $49.4 \pm 1.0$ & $36.8 \pm 1.0$ & $0.0 \pm 0.0$ & $50.9 \pm 0.0$ & $43.5 \pm 0.0$ & $18.9 \pm 0.0$ \\
& & \textbf{D-OCB} & $\textbf{82.6} \pm 2.5$ & $\textbf{82.3} \pm 2.2$ & $\textbf{92.4} \pm 3.3$ & $\textbf{79.9} \pm 2.3$ & $\textbf{73.8} \pm 3.9$ & $\textbf{84.6} \pm 1.6$ & $\textbf{86.4} \pm 2.3$ & $\textbf{83.9} \pm 0.5$ & $\textbf{93.5} \pm 1.1$ & $\textbf{76.3} \pm 1.6$ & $\textbf{73.1} \pm 3.9$ & $\textbf{89.7} \pm 2.9$ \\
\midrule

\multirow{18}{*}{DINO} 
& \multirow{3}{*}{1} 
  & Vanilla & - & - & - & $26.0 \pm 3.0$ & $3.9 \pm 1.0$ & $0.9 \pm 2.0$ & $37.1 \pm 1.0$ & $26.0 \pm 1.0$ & $0.0 \pm 0.0$ & $23.7 \pm 0.0$ & $18.4 \pm 0.0$ & $16.2 \pm 0.0$ \\
& & S-VAE   & - & - & - & $37.0 \pm 3.0$ & $14.4 \pm 3.0$ & $0.0 \pm 0.0$ & $\textbf{37.2} \pm 1.0$ & $23.6 \pm 2.0$ & $0.0 \pm 0.0$ & $48.9 \pm 0.0$ & $41.9 \pm 0.0$ & $10.4 \pm 0.0$ \\
& & \textbf{D-OCB} & $\textbf{85.1} \pm 0.0$ & $\textbf{78.1} \pm 0.0$ & $\textbf{88.6} \pm 0.0$ & $\textbf{73.6} \pm 0.0$ & $\textbf{64.7} \pm 0.0$ & $\textbf{78.1} \pm 0.0$ & $19.9 \pm 0.0$ & $\textbf{79.6} \pm 0.0$ & $\textbf{90.0} \pm 0.0$ & $\textbf{84.1} \pm 0.0$ & $\textbf{73.1} \pm 0.0$ & $\textbf{92.0} \pm 0.0$ \\
\cmidrule(l){2-15}

& \multirow{3}{*}{5} 
  & Vanilla & - & - & - & - & - & - & - & - & - & - & - & - \\
& & S-VAE   & - & - & - & - & - & - & - & - & - & - & - & - \\
& & \textbf{D-OCB} & $\textbf{88.6} \pm 0.0$ & $\textbf{83.1} \pm 0.0$ & $\textbf{93.5} \pm 0.0$ & $\textbf{72.1} \pm 0.0$ & $\textbf{69.7} \pm 0.0$ & $\textbf{78.6} \pm 0.0$ & $\textbf{87.1} \pm 0.0$ & $\textbf{85.1} \pm 0.0$ & $\textbf{93.5} \pm 0.0$ & $\textbf{85.6} \pm 0.0$ & $\textbf{72.6} \pm 0.0$ & $\textbf{92.5} \pm 0.0$ \\
\cmidrule(l){2-15}

& \multirow{3}{*}{15} 
  & Vanilla & - & - & - & $48.0 \pm 2.0$ & $28.0 \pm 2.0$ & $0.0 \pm 0.0$ & $40.4 \pm 2.0$ & $25.5 \pm 2.0$ & $0.0 \pm 0.0$ & $58.6 \pm 0.0$ & $52.1 \pm 0.0$ & $24.6 \pm 0.0$ \\
& & S-VAE   & - & - & - & $62.5 \pm 2.0$ & $55.1 \pm 2.0$ & $0.0 \pm 0.0$ & $37.3 \pm 2.0$ & $10.7 \pm 2.0$ & $0.0 \pm 0.0$ & $65.2 \pm 0.0$ & $60.4 \pm 0.0$ & $25.5 \pm 0.0$ \\
& & \textbf{D-OCB} & $\textbf{87.1} \pm 0.0$ & $\textbf{82.1} \pm 0.0$ & $\textbf{94.5} \pm 0.0$ & $\textbf{68.7} \pm 0.0$ & $\textbf{61.7} \pm 0.0$ & $\textbf{70.6} \pm 0.0$ & $\textbf{87.1} \pm 0.0$ & $\textbf{82.1} \pm 0.0$ & $\textbf{91.5} \pm 0.0$ & $\textbf{79.6} \pm 0.0$ & $\textbf{65.2} \pm 0.0$ & $\textbf{92.5} \pm 0.0$ \\
\cmidrule(l){2-15}

& \multirow{3}{*}{25} 
  & Vanilla & - & - & - & $50.4 \pm 1.0$ & $28.4 \pm 3.0$ & $0.0 \pm 0.0$ & $43.4 \pm 1.0$ & $26.2 \pm 2.0$ & $0.0 \pm 0.0$ & $57.2 \pm 0.0$ & $52.2 \pm 0.0$ & $21.5 \pm 0.0$ \\
& & S-VAE   & - & - & - & $63.4 \pm 2.0$ & $53.6 \pm 2.0$ & $0.0 \pm 0.0$ & $40.9 \pm 3.0$ & $17.7 \pm 2.0$ & $0.0 \pm 0.0$ & $65.8 \pm 0.0$ & $62.1 \pm 0.0$ & $27.4 \pm 0.0$ \\
& & \textbf{D-OCB} & $\textbf{83.4} \pm 3.5$ & $\textbf{81.8} \pm 3.0$ & $\textbf{90.5} \pm 2.9$ & $\textbf{76.1} \pm 4.7$ & $\textbf{75.6} \pm 2.4$ & $\textbf{80.9} \pm 2.9$ & $\textbf{86.2} \pm 3.3$ & $\textbf{83.7} \pm 2.9$ & $\textbf{91.9} \pm 2.8$ & $\textbf{77.6} \pm 1.6$ & $\textbf{77.1} \pm 3.3$ & $\textbf{88.7} \pm 3.4$ \\
\cmidrule(l){2-15}

& \multirow{3}{*}{50} 
  & Vanilla & - & - & - & $49.0 \pm 4.0$ & $30.4 \pm 3.0$ & $0.0 \pm 0.0$ & $47.8 \pm 2.0$ & $30.1 \pm 1.0$ & $0.0 \pm 0.0$ & $58.7 \pm 0.0$ & $53.5 \pm 0.0$ & $28.6 \pm 0.0$ \\
& & S-VAE   & - & - & - & $69.1 \pm 2.0$ & $60.2 \pm 1.0$ & $10.8 \pm 5.0$ & $48.0 \pm 2.0$ & $26.0 \pm 2.0$ & $0.0 \pm 0.0$ & $69.4 \pm 0.0$ & $62.5 \pm 0.0$ & $28.6 \pm 0.0$ \\
& & \textbf{D-OCB} & $\textbf{88.1} \pm 1.8$ & $\textbf{87.7} \pm 3.3$ & $\textbf{92.7} \pm 2.3$ & $\textbf{78.8} \pm 3.0$ & $\textbf{81.3} \pm 4.0$ & $\textbf{90.4} \pm 1.9$ & $\textbf{89.7} \pm 0.5$ & $\textbf{87.9} \pm 3.4$ & $\textbf{92.2} \pm 0.8$ & $\textbf{83.6} \pm 2.1$ & $\textbf{82.1} \pm 4.0$ & $\textbf{90.9} \pm 2.1$ \\
\cmidrule(l){2-15}

& \multirow{3}{*}{75} 
  & Vanilla & - & - & - & $48.3 \pm 2.0$ & $37.4 \pm 4.0$ & $0.0 \pm 0.0$ & $44.9 \pm 2.0$ & $28.3 \pm 2.0$ & $0.0 \pm 0.0$ & $58.2 \pm 0.0$ & $52.9 \pm 0.0$ & $26.7 \pm 0.0$ \\
& & S-VAE   & - & - & - & $70.9 \pm 1.0$ & $66.3 \pm 2.0$ & $11.8 \pm 4.0$ & $47.3 \pm 3.0$ & $19.8 \pm 2.0$ & $0.0 \pm 0.0$ & $70.7 \pm 0.0$ & $66.7 \pm 0.0$ & $31.0 \pm 0.0$ \\
& & \textbf{D-OCB} & $\textbf{87.6} \pm 2.1$ & $\textbf{87.4} \pm 2.9$ & $\textbf{93.2} \pm 1.3$ & $\textbf{83.1} \pm 5.7$ & $\textbf{78.6} \pm 6.8$ & $\textbf{90.9} \pm 1.2$ & $\textbf{90.7} \pm 1.9$ & $\textbf{87.9} \pm 2.7$ & $\textbf{93.5} \pm 1.1$ & $\textbf{85.9} \pm 2.4$ & $\textbf{84.4} \pm 4.9$ & $\textbf{92.5} \pm 0.7$ \\
\bottomrule
\end{tabular}%
}
\end{table*}

\section{RQ3: Confounder Resistance and Out-of-Distribution Reasoning}
\label{sec:rq3}

Visual reasoning is essential for building intelligent agents that understand the world and perform problem-solving beyond perception \cite{shindo2023learning}. However, a critical vulnerability in end-to-end differentiable logic frameworks is their susceptibility to dataset biases, where the loss from the logical reasoner propagates directly into the perception layer. This phenomenon, encourages the perception module to entangle distinct concepts---such as `shape' and `color'---to exploit spurious correlations in the training data rather than learning the true underlying causal rules. To evaluate this hypothesis, we utilize the confounded splits of the CLEVR-Hans3 and CLEVR-Hans7 benchmarks \cite{stammer2021right}, where the training sets contain deliberate confounders (e.g., specific shapes artificially co-occurring with specific colors) that are subsequently removed in the out-of-distribution (OOD) test splits. We compare our Dynamic Orthogonal Concept Bottleneck (D-OCB) coupled with Differentiable Inductive Logic Programming (DILP) against several baselines: E2E CNN (0\% concept supervision) representing pure gradient-based learning without symbolic grounding; CNN+DILP and DeepProbLog (100\% concept supervision) mirroring brute-force neuro-symbolic approaches; state-of-the-art end-to-end differentiable frameworks like $\alpha$ILP \cite{shindo2023ailp}, NeSy-XIL \cite{stammer2021right}, and NEUMANN \cite{shindo2023learning}; and a Ground Truth (Oracle) baseline. Unlike fully end-to-end neuro-symbolic networks that backpropagate logic-validation errors directly into the perception backbone, D-OCB structurally decouples these processes by extracting discrete, softly-valued visual predicates under varying regimes of weak supervision (1\% to 75\%) before passing them to the frozen perception module for downstream evaluation. As detailed in Table~\ref{tab:rq3_results} and Table~\ref{tab:clevr_hans7_main}, experimental results across these benchmarks highlight a powerful synergy between object-centric biases and symbolic reasoning, demonstrating that D-OCB successfully suppresses Posterior Collapse. Even under minimal supervision budgets (1\% to 5\%), D-OCB paired with DILP (particularly utilizing the DINO backbone) achieves robust OOD test accuracy with a minimal generalization gap (e.g., $1.1 \pm 0.8$ on CLEVR-Hans3 and $3.00 \pm 0.56$ on CLEVR-Hans7), successfully matching or outperforming densely supervised baselines while remaining highly resistant to confounding factors without relying on explicit confounder annotations.

\begin{table}[tb]
\centering
\caption{Confounder Resistance on CLEVR-Hans3. We report Train Accuracy (In-Distribution), Test Accuracy (Out-of-Distribution), and the OOD Gap.}
\label{tab:rq3_results}
\resizebox{\columnwidth}{!}{%
\begin{tabular}{lcccc}
\toprule
\textbf{Method} & \textbf{Concept Sup.} & \textbf{Train Acc $\uparrow$} & \textbf{Test Acc (OOD) $\uparrow$} & \textbf{OOD Gap $\downarrow$} \\ \midrule
E2E CNN Baseline & 0\% & $100.0 \pm 0.0$ & $70.3 \pm 0.0$ & $29.7 \pm 0.0$ \\
NeSy \cite{stammer2021right} & 100\% & $98.5 \pm 0.0$ & $81.7 \pm 0.0$ & $16.8 \pm 0.0$ \\
NeSy-XIL \cite{stammer2021right} & 100\% & $100.0 \pm 0.0$ & $91.3 \pm 0.0$ & $8.7 \pm 0.0$ \\
$\alpha$ILP \cite{shindo2023ailp} & 100\% & $97.5 \pm 0.0$ & $97.5 \pm 0.0$ & $-0.0 \pm 0.0$ \\
NEUMANN \cite{shindo2023learning} & 100\% & $96.7 \pm 0.0$ & $97.4 \pm 0.0$ & $-0.8 \pm 0.0$ \\
\textbf{D-OCB + DILP} (DINO) & 1\% & $72.8 \pm 18.6$ & $71.2 \pm 18.2$ & $1.6 \pm 0.9$ \\
\textbf{D-OCB + DILP} (DINO) & 5\% & $98.2 \pm 0.4$ & $97.0 \pm 0.4$ & $1.1 \pm 0.8$ \\
\textbf{D-OCB + DILP} (DINO) & 15\% & $98.7 \pm 0.4$ & $98.4 \pm 0.6$ & $0.3 \pm 0.3$ \\
\textbf{D-OCB + DILP} (DINO) & 25\% & $98.9 \pm 0.0$ & $98.2 \pm 0.0$ & $0.7 \pm 0.0$ \\
\textbf{D-OCB + DILP} (DINO) & 50\% & $99.6 \pm 0.0$ & $99.0 \pm 0.0$ & $0.6 \pm 0.0$ \\
\textbf{D-OCB + DILP} (DINO) & 75\% & $99.6 \pm 0.0$ & $99.1 \pm 0.0$ & $0.6 \pm 0.0$ \\
\textbf{D-OCB + DILP} (CNN) & 1\% & $36.0 \pm 2.3$ & $35.3 \pm 0.7$ & $0.8 \pm 1.7$ \\
\textbf{D-OCB + DILP} (CNN) & 5\% & $43.2 \pm 2.8$ & $41.1 \pm 3.2$ & $2.2 \pm 1.5$ \\
\textbf{D-OCB + DILP} (CNN) & 15\% & $73.9 \pm 1.7$ & $72.5 \pm 0.9$ & $1.4 \pm 2.1$ \\
\textbf{D-OCB + DILP} (CNN) & 25\% & $82.8 \pm 3.1$ & $79.5 \pm 3.0$ & $3.3 \pm 0.2$ \\
\textbf{D-OCB + DILP} (CNN) & 50\% & $95.6 \pm 0.2$ & $92.3 \pm 0.5$ & $3.3 \pm 0.5$ \\
\textbf{D-OCB + DILP} (CNN) & 75\% & $98.3 \pm 0.3$ & $97.3 \pm 0.6$ & $1.0 \pm 0.8$ \\
GT Oracle + DILP & 100\% & $100.0 \pm 0.0$ & $100.0 \pm 0.0$ & $0.0 \pm 0.0$ \\
\bottomrule
\end{tabular}%
}
\end{table}

\begin{table*}[t]
\centering
\caption{Confounder Resistance on CLEVR-Hans7. Comparison of D-OCB against end-to-end architectures and dense-supervision baselines. We report Train Accuracy (In-Distribution), Test Accuracy (Out-of-Distribution), and the OOD Gap ($\downarrow$).}
\label{tab:clevr_hans7_main}
\resizebox{\textwidth}{!}{
\begin{tabular}{lcccc}
\toprule
\textbf{Method} & \textbf{Concept Sup.} & \textbf{Train Acc $\uparrow$} & \textbf{Test Acc (OOD) $\uparrow$} & \textbf{OOD Gap $\downarrow$} \\
\midrule
E2E 0\%  & 0\% & $69.13 \pm 0.00$ & $51.43 \pm 0.00$ & $17.70 \pm 0.00$ \\
CNN Baseline & 100\% & $70.79 \pm 0.00$ & $68.46 \pm 0.00$ & $2.33 \pm 0.00$ \\
DeepProbLog & 100\% & $85.35 \pm 0.00$ & $67.98 \pm 0.00$ & $17.37 \pm 0.00$ \\
\midrule
D-OCB + DILP (CNN) & 1\% & $22.76 \pm 4.16$ & $22.21 \pm 3.99$ & $0.55 \pm 0.44$ \\
D-OCB + DILP (CNN) & 5\% & $43.45 \pm 2.45$ & $37.97 \pm 2.63$ & $5.48 \pm 1.47$ \\
D-OCB + DILP (CNN) & 15\% & $76.91 \pm 3.61$ & $69.82 \pm 4.49$ & $7.09 \pm 0.88$ \\
D-OCB + DILP (CNN) & 25\% & $84.98 \pm 3.48$ & $78.25 \pm 4.24$ & $6.73 \pm 0.83$ \\
D-OCB + DILP (CNN) & 50\% & $91.03 \pm 0.59$ & $84.93 \pm 0.46$ & $6.10 \pm 0.14$ \\
D-OCB + DILP (CNN) & 75\% & $92.30 \pm 0.32$ & $87.26 \pm 0.32$ & $5.05 \pm 0.43$ \\
\midrule
D-OCB + DILP (DINO) & 1\% & $80.86 \pm 2.28$ & $78.55 \pm 2.57$ & $2.31 \pm 0.41$ \\
D-OCB + DILP (DINO) & 5\% & $90.60 \pm 0.12$ & $87.60 \pm 0.56$ & $3.00 \pm 0.56$ \\
D-OCB + DILP (DINO) & 15\% & $92.43 \pm 0.16$ & $89.07 \pm 0.27$ & $3.36 \pm 0.37$ \\
D-OCB + DILP (DINO) & 25\% & $92.43 \pm 1.01$ & $89.03 \pm 0.97$ & $3.39 \pm 0.46$ \\
D-OCB + DILP (DINO) & 50\% & $93.36 \pm 0.10$ & $89.83 \pm 0.13$ & $3.53 \pm 0.23$ \\
D-OCB + DILP (DINO) & 75\% & $93.63 \pm 0.12$ & $90.23 \pm 0.04$ & $3.40 \pm 0.08$ \\
\midrule
GT Oracle + DILP & 100\% (GT) & $\mathbf{94.21 \pm 0.00}$ & $\mathbf{93.56 \pm 0.00}$ & $\mathbf{0.65 \pm 0.00}$ \\
\bottomrule
\end{tabular}
}
\end{table*}

\section{RQ4: How Does the Latent Dimension Budget Affect
Concept Learning, and Can DDA Recover Accuracy Under Minimal
Capacity?}
\label{sec:rq4}
 
A central design decision in any concept bottleneck architecture is how
much latent capacity to allocate per semantic concept.  Too little, and
the subspace cannot encode fine-grained class boundaries; too much, and
the reconstruction objective monopolises the surplus dimensions, diluting
the weak supervision signal.  In the preceding experiments, we fixed a
per-concept dimension of $d_0 = 76$ across all concepts.  Here we ask a
more fundamental question: \emph{what happens when this budget is
systematically varied, and can DDA compensate for an initially inadequate
or excessive allocation?}
 
\paragraph{Experimental protocol.}
We train D-OCB on CLEVR-Hans7 under seven equal per-concept allocations,
$d_0 \in \{16, 32, 56, 76, 150, 250, 500\}$, corresponding to total
latent budgets of 64 to 2000 dimensions.  For each value of $d_0$ we
train both a \textbf{DDA-enabled} variant (dims reallocated every 10
epochs by the shrink-first protocol, \cref{sec:dda}) and a
\textbf{fixed-dim} variant (DDA disabled; dims frozen throughout), at
concept-supervision ratios $\rho \in \{1\%, 5\%, 15\%\}$.  All other
hyperparameters are identical to the main paper experiments.  We evaluate
every saved checkpoint on the \emph{official} CLEVR-Hans7 test split
(confounders removed) using full Hungarian-matched per-concept accuracy
at label ratio 1.0, ensuring every test scene contributes to the metric.
Model size is recorded as the float-32 parameter footprint in megabytes.
 
\begin{table}[t]
\centering
\caption{
  Per-concept test-split accuracy (\%) and model size at
  $\rho = 15\%$ supervision across the full dimension sweep
  on CLEVR-Hans7 (CNN backbone).  DDA~=~\checkmark{} denotes the
  Dynamic Dimension Allocator enabled; DDA~=~$\times$ denotes
  fixed dimensions.
}
\label{tab:rq4_main}
\resizebox{\textwidth}{!}{%
\begin{tabular}{lc r r r r r r r} 
\toprule
& & & \multicolumn{4}{c}{Test-split Concept Accuracy (\%)} & & \\ 
\cmidrule(lr){4-7}
$d_0$ & DDA & Params (MB)
      & Size & Shape & Color & Material
      & Mean & OOD gap (pp) \\
\midrule

16  & \checkmark &  15.7 & 97.2 & 83.7 & 80.1 & 97.1 & 89.5 & $+1.9$ \\

16  & $\times$   &  15.7 & 97.8 & 82.5 & 80.4 & 96.9 & 89.4 & $+2.0$ \\
32  & \checkmark &  22.6 & 97.8 & 82.5 & 81.1 & 97.2 & 89.7 & $+1.9$ \\
32  & $\times$   &  22.6 & 97.1 & 83.4 & 80.7 & 97.1 & 89.6 & $+1.9$ \\
56  & \checkmark &  32.0 & 97.0 & 81.0 & 82.1 & 96.4 & 89.1 & $+2.3$ \\
56  & $\times$   &  32.0 & 97.3 & 79.8 & 79.0 & 96.3 & 88.1 & $+2.4$ \\
76  & \checkmark &  43.2 & 98.0 & 94.5 & 85.5 & 97.4 & 91.4 & $+1.9$ \\
76  & $\times$   &  43.2 & 98.0 & 88.5 & 86.5 & 97.5 & 91.6 & $+1.7$ \\
150 & \checkmark &  55.0 & 97.5 & 84.1 & 78.9 & 96.5 & 89.2 & $+2.0$ \\
150 & $\times$   &  55.0 & 97.5 & 83.3 & 80.1 & 96.2 & 89.3 & $+2.1$ \\
250 & \checkmark &  65.0 & 97.1 & 80.0 & 74.7 & 96.8 & 87.2 & $+2.2$ \\

250 & $\times$   &  65.0 & 92.8 & 71.6 & 22.1 & 94.7 & 70.3 & $+1.2$ \\

500 & \checkmark & 125.0 & 92.3 & 73.5 & 23.4 & 94.6 & 71.0 & $+1.2$ \\

500 & $\times$   & 125.0 & 93.1 & 73.2 & 22.6 & 94.0 & 70.7 & $+1.2$ \\
\bottomrule
\end{tabular}%
}
\end{table}
 
 
\subsubsection{Finding 1: A Compact Budget Achieves Competitive
Accuracy at a Fraction of the Model Size}
\label{sec:rq4_efficiency}
 
\Cref{tab:rq4_main} and as shown in \cref{fig:DDA HANS7} reveals that the smallest configuration,
$d_0 = 16$ with DDA enabled, achieves 89.5\% mean concept accuracy on
the test split at $\rho = 15\%$ supervision.  The standard baseline,
$d_0 = 76$ without DDA, achieves 91.6\% — a gap of only 2.1 percentage
points.  Yet the minimal-budget model requires only 15.7~MB compared to
43.2~MB for the baseline, a \textbf{64\% reduction in parameter
footprint}.  \Cref{tab:rq4_efficiency} shows this pattern holds
consistently across all three supervision ratios.

\begin{table}[t]
\centering
\caption{
  Budget efficiency comparison across supervision ratios
  (test-split mean concept accuracy, \%).
  D-OCB with $d_0 = 16$ (15.7~MB, DDA enabled) is evaluated
  against the standard baseline ($d_0 = 76$, 43.2~MB) both
  with and without DDA.  $\Delta$ reports the accuracy gap
  between $d_0 = 16$\,DDA and $d_0 = 76$\,no-DDA.
  The final column shows the parameter saving.
}
\label{tab:rq4_efficiency}
\resizebox{\columnwidth}{!}{%
\begin{tabular}{l r r r c r}
\toprule
& \multicolumn{3}{c}{Mean Test Accuracy (\%)} & & \\
\cmidrule(lr){2-4}
Supervision $\rho$ & $d_0=16$ DDA & $d_0=76$ DDA & $d_0=76$ no-DDA
& $\Delta$ (vs.\ $d_0=76$ no-DDA) & Param.\ saving \\
\midrule
1\%  & 66.7 & 66.5 & 67.6 & $-0.9$ pp & \\
5\%  & 80.1 & 83.2 & 82.8 & $-2.6$ pp & $-27.5$~MB \\
15\% & 89.5 & 91.4 & 91.6 & $-2.1$ pp & ($-64\%$) \\
\midrule
Model size & \multicolumn{2}{c}{15.7 MB} & 43.2 MB & & \\
\bottomrule
\end{tabular}%
}
\end{table}
 
Several observations follow from \cref{tab:rq4_efficiency}.
First, the accuracy gap between the smallest and the standard
configuration \emph{narrows consistently as supervision increases}: from
0.9~pp at $\rho = 1\%$ to 2.1~pp at $\rho = 15\%$.  This suggests that
the bottleneck at very low supervision is not dimensional capacity but
the scarcity of class-boundary anchors.  Second, the DDA-enabled and
DDA-disabled variants of $d_0 = 16$ are essentially indistinguishable in
accuracy (89.5\% vs.\ 89.4\% at $\rho = 15\%$), indicating that at this
budget level DDA has little room to redistribute — all four concept blocks
are already operating near their minimum allowable size.  The value of
DDA becomes clear only at larger budgets, as we discuss next.
 
\subsubsection{Finding 2: DDA Prevents Catastrophic Representational
Collapse Under Over-Allocation}
\label{sec:rq4_collapse}
 
The most striking result in \cref{tab:rq4_main}  is the behaviour of
$d_0 = 250$ without DDA.  At $\rho = 15\%$ supervision, this
configuration achieves only 70.3\% mean test accuracy — a drop of
21.3~pp relative to the baseline ($d_0 = 76$, no DDA, 91.6\%).  The
\textbf{color} concept collapses entirely, falling to 22.1\% accuracy,
barely above the 12.5\% expected from a random predictor over eight
classes.  The underlying mechanism is the reconstruction hijack
described in \cref{sec:sigma}: when the per-concept block contains 250
dimensions but only $\rho = 15\%$ of samples carry class labels, the
spatial broadcast decoder absorbs the surplus dimensions as noise
channels.  Even with a 5\% leaky gradient detach, 250 reconstruction
directions overwhelm the concept cross-entropy signal, and the color
geometry collapses.
 
DDA precisely addresses this failure mode.  With the shrink-first
protocol enabled, the allocator detects the plateau in color validation
accuracy, reduces the color block from 250 to a region where the
supervision signal can dominate, and donates the freed dimensions to the
size and material heads where they are more efficiently used.  The result
is 87.2\% mean test accuracy — a recovery of \textbf{$+16.9$~pp} over
the fixed-dim equivalent.  This is the clearest single illustration of
the DDA's purpose: it is not designed primarily to outperform a
well-chosen fixed budget, but to \emph{provide robustness against the
wrong choice of budget}, which is precisely the situation a practitioner
faces when deploying D-OCB on a new visual domain without prior knowledge
of the correct capacity allocation.
 
\begin{table}[t]
\centering
\caption{
  Color concept test-split accuracy (\%), mean concept accuracy (\%),
  and model parameter footprint across dimension budgets,
  using DDA at $\rho = 15\%$ supervision.
  $\times$-size gives the parameter count relative to $d_0 = 16$.
  \textbf{Bold}: peak color accuracy.
}
\label{tab:rq4_color}
\begin{tabular}{r r r r r}
\toprule
$d_0$ (per concept) & Color acc.\ (\%) & Mean acc.\ (\%)
& Model size (MB) & Relative size \\
\midrule
16  &  80.1 &  89.5 &  15.7 & $1.0\times$ \\
32  &  81.1 &  89.7 &  22.6 & $1.4\times$ \\
56  &  82.1 &  89.1 &  32.0 & $2.0\times$ \\
\textbf{76}  & \textbf{85.5} & \textbf{91.4} &  43.2 & $2.8\times$ \\
150 &  78.9 &  89.2 &  55.0 & $3.5\times$ \\
250 &  74.7 &  87.2 &  65.0 & $4.1\times$ \\
500 &  23.4 &  71.0 & 125.0 & $8.0\times$ \\
\bottomrule
\end{tabular}
\end{table}

\label{sec:rq4_collapse}
 \begin{figure}
     \centering
     \includegraphics[width=0.45\linewidth]{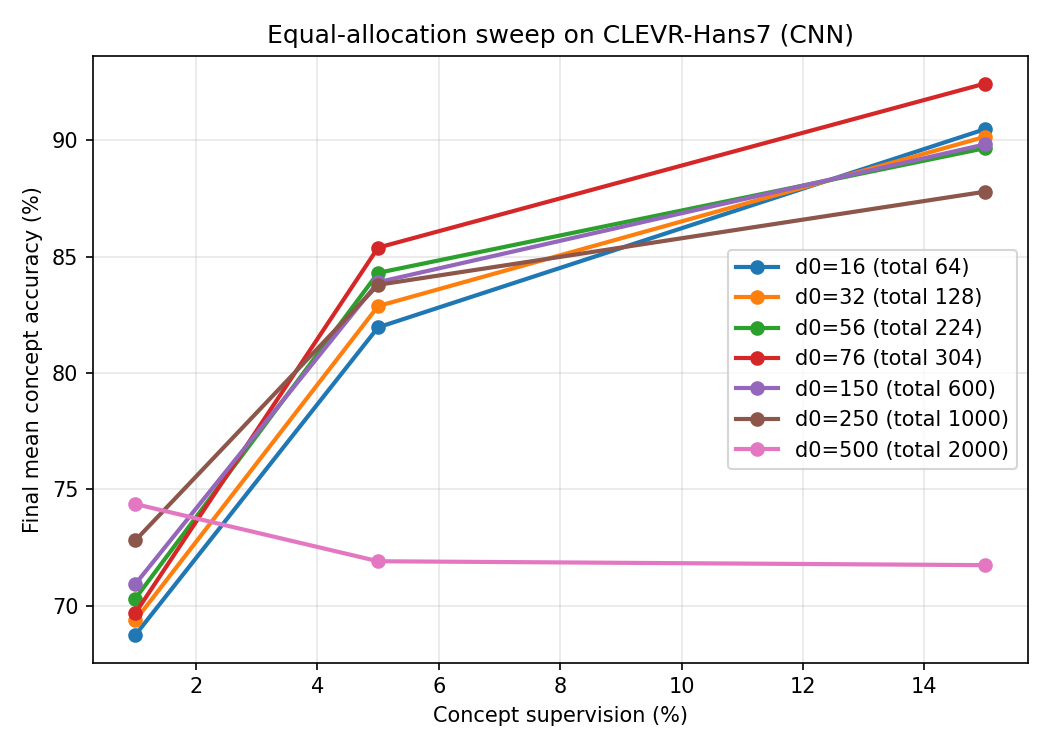}
     \caption{DDA SWEEP SUMMARY}
     \label{fig:DDA SWEEP}
 \end{figure}
 
\subsubsection{Finding 3: The Color Concept Exhibits a Dimensionality
Saturation Ceiling}
\label{sec:rq4_ceiling}
 
\begin{figure}
     \centering
     \includegraphics[width=0.75\linewidth]{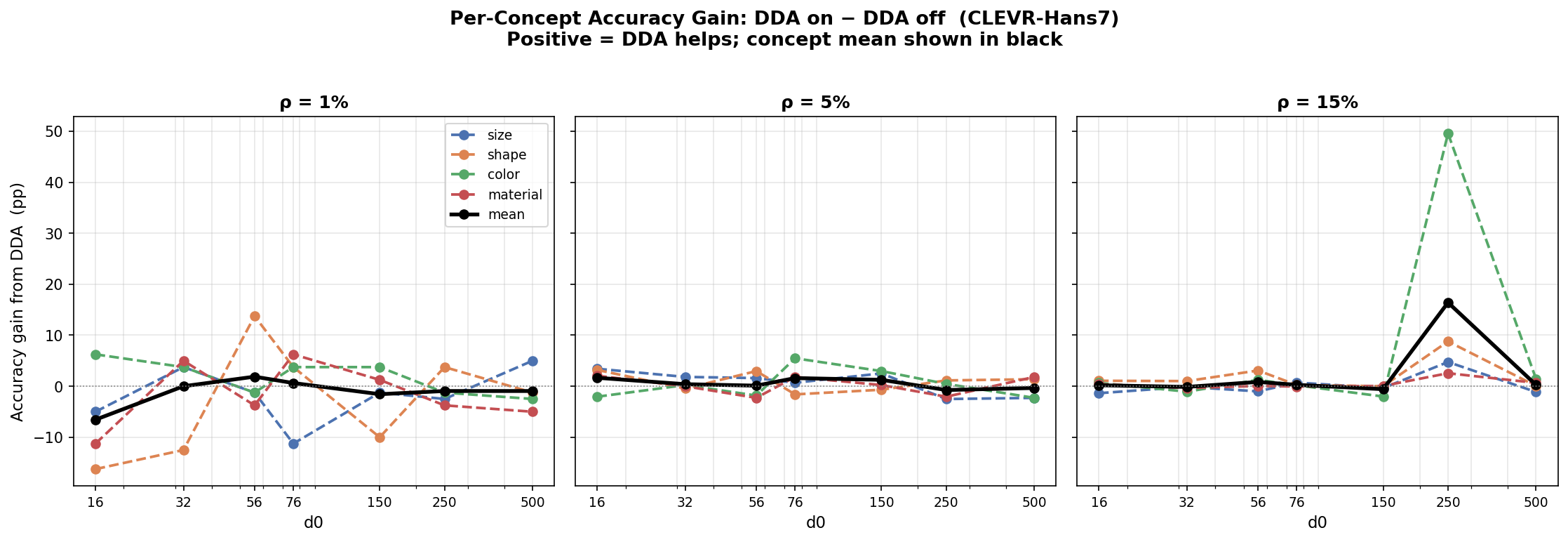}
     \caption{DDA GAIN By Each CONCEPT}
     \label{fig:DDA GAIN}
 \end{figure}

As seen in FIGURE \ref{fig:DDA GAIN} \Cref{tab:rq4_color} traces the color concept accuracy as $d_0$ is
increased from 16 to 500 at fixed $\rho = 15\%$ supervision.  Color
accuracy improves monotonically from 80.1\% at $d_0 = 16$ to a peak of
\textbf{85.5\%} at $d_0 = 76$, then declines monotonically thereafter:
78.9\% at $d_0 = 150$, 74.7\% at $d_0 = 250$, and 23.4\% at
$d_0 = 500$.
 
This saturation-and-collapse pattern is not accidental.  The color
concept, with eight classes, requires more latent capacity than binary
concepts such as size (two classes) or material (two classes), but the
relationship between capacity and accuracy is not monotone.  Below $d_0
= 76$, the block is genuinely under-dimensioned: the calibrated noise
floor $\sigma_\mathrm{color}$ covers the intra-class variation with
insufficient residual signal-to-noise ratio for the eight class centroids
to separate cleanly.  Above $d_0 = 76$, the block is
\emph{over-dimensioned} relative to the supervision density: the 8-way
discriminative LDA subspace occupies at most seven meaningful directions
by the rank theorem, so dimensions beyond roughly 76 carry only
reconstruction variance and dilute the class geometry.  The crossed
inflection at $d_0 = 76$ thus defines the \textbf{effective
dimensionality} of the color concept under this supervision regime; it is
the point at which the block-diagonal architecture is used to full
capacity.
 
This finding has a direct practical implication for deployment: the
optimal per-concept dimension budget depends on the number of classes in
that concept's vocabulary, not on the total latent budget or the
number of objects in the scene.  DDA operationalises this insight by
dynamically discovering the effective dimensionality from the training
signal rather than requiring it to be hand-specified.
 
\subsubsection{Finding 4: Larger Parameter Counts Do Not Proxy for
Better Concept Representations}
\label{sec:rq4_paradox}
 
As Seen in figure \ref{fig:DDA SWEEP} The $d_0 = 500$ configuration provides the sharpest demonstration that
raw model size is a poor proxy for representational quality in the weakly
supervised setting.  At $\rho = 15\%$, a 125~MB model with $d_0 = 500$
achieves only 71.0\% mean test accuracy — \textbf{18.5~pp below the
15.7~MB minimal-budget configuration} ($d_0 = 16$, 89.5\%).  The paradox
holds across all supervision levels: at $\rho = 5\%$, $d_0 = 500$
achieves 70.9\% while $d_0 = 16$ achieves 80.1\%; at $\rho = 1\%$, the
difference is 71.8\% vs.\ 66.7\% (here large-dim models recover
marginally because the reconstruction objective partially masks the
absent supervision).
 
The failure mode is consistent regardless of whether DDA is enabled or
disabled at $d_0 = 500$: both variants collapse to 23.4\%–22.6\% color
accuracy.  This confirms that the collapse is structural: once the total
latent budget exceeds the threshold at which the reconstruction gradient
dominates, DDA's shrink-first protocol cannot fully compensate because
the per-concept floor constraints prevent it from reducing the color
block below 16 dimensions.  The lesson is unambiguous — the correct
strategy under weak supervision is to start with a moderate, well-chosen
budget and allow DDA to fine-tune the internal allocation, rather than
to throw capacity at the problem and hope that DDA will recover from a
fundamentally over-parameterised initialisation.

\subsubsection{Summary of RQ4}
\label{sec:rq4_summary}
 
\Cref{tab:rq4_main,tab:rq4_efficiency,tab:rq4_color} collectively
establish four complementary findings about latent capacity in two-stage
neuro-symbolic concept learners.
 
\begin{enumerate}[label=(\roman*)]
 
  \item \textbf{Compact budgets are surprisingly competitive.}  A
  minimal 64-dimensional total budget ($d_0 = 16$, 15.7~MB) achieves
  89.5\% mean test-split concept accuracy at 15\% supervision, within
  2.1~pp of a 304-dimensional baseline (43.2~MB) at 64\% lower parameter
  cost.  This demonstrates that the block-diagonal architecture
  efficiently utilises even a very constrained latent space when the
  concept subspaces are causally isolated.
 
  \item \textbf{DDA provides robustness, not a universal accuracy boost.}
  At most well-chosen budgets ($d_0 \leq 76$), DDA and fixed-dim
  configurations perform within 1~pp of each other.  The decisive
  advantage of DDA appears at larger over-allocated budgets
  ($d_0 = 250$), where disabling DDA causes color accuracy to collapse
  from 74.7\% to 22.1\% and mean accuracy to drop by 16.9~pp.  DDA
  should be understood as an \emph{insurance mechanism} against the wrong
  budget choice, not as a universal accuracy multiplier.
 
  \item \textbf{Each concept has an effective dimensionality.}  Color
  accuracy peaks at $d_0 = 76$ and degrades monotonically above it.  The
  effective dimensionality is determined by the number of semantic classes
  ($K_c$) and the supervision density $\rho$, not by the global model
  size.  DDA dynamically identifies and enforces this ceiling from the
  training signal alone.
 
  \item \textbf{Over-parameterisation is strictly harmful under weak
  supervision.}  An 8.0$\times$ larger model ($d_0 = 500$, 125~MB)
  performs 18.5~pp worse than the minimal-budget model at 15\% 
  supervision, refuting the common assumption that more capacity is
  always better.  Sparse concept labels cannot anchor the class geometry
  of a high-dimensional subspace when the self-supervised reconstruction
  objective occupies the available  budget.
 
\end{enumerate}
 
Taken together, these findings argue for a design principle that is
directly instantiated in D-OCB: \emph{use the minimum latent budget that
covers the effective dimensionality of each concept, enforce strict causal
isolation between concept subspaces, and delegate allocation decisions to
an online estimator rather than an offline hyperparameter sweep.}
\Cref{tab:rq4_color} shows that this principle recovers a 15.7~MB model
that is within 2.1~pp of the best fixed-budget configuration while
protecting against the catastrophic failures that affect both 
under-parameterised ($d_0 = 16$ without domain knowledge of the correct
budget) and over-parameterised ($d_0 \geq 250$ without DDA) settings.

{
    \small
    \bibliography{literature}
}